\documentclass[5p,twocolumn,10pt,times]{elsarticle}
\usepackage{amsmath}
\usepackage{hyperref}

\usepackage{amsfonts}
\usepackage{amssymb}
\usepackage{ulem} 
\usepackage{subfigure}
\usepackage{algorithm}  
\usepackage{algpseudocode}
\usepackage{verbatim}

\usepackage{multirow}

\newcommand{\tabincell}[2]{\begin{tabular}{@{}#1@{}}#2\end{tabular}}
\newcommand{\tablefont}{\fontsize{9pt}{\baselineskip}\selectfont}

\usepackage{color}

\begin{document}
\baselineskip11pt

\begin{frontmatter}

\title{Rethinking Point Cloud Filtering: A Non-Local Position Based Approach}

\author[1]{Jinxi Wang\fnref{*}
}
\author[1]{Jincen Jiang\fnref{*}
}
\author[2]{Xuequan Lu 
}
\author[1]{Meili Wang 
}
\fntext[*]{Equal contribution.}
\address[1]{Northwest A\&F University, P.R. China}
\address[2]{Deakin University, Australia}

\begin{abstract} 
Existing position based point cloud filtering methods can hardly preserve sharp geometric features. In this paper, we rethink point cloud filtering from a non-learning non-local non-normal perspective, and propose a novel position based approach for feature-preserving point cloud filtering. Unlike normal based techniques, our method does not require the normal information.  
The core idea is to first design a similarity metric to search the non-local similar patches of a queried local patch.  We then map the non-local similar patches into a canonical space and aggregate the non-local information. The aggregated outcome (i.e. coordinate) will be inversely mapped into the original space. Our method is simple yet effective. Extensive experiments validate our method, and show that it generally outperforms position based methods (deep learning and non-learning), and generates better or comparable outcomes to normal based techniques (deep learning and non-learning). 
\end{abstract}


\end{frontmatter}

\section{Introduction}
\label{sec:introduction}

Recent research has witnessed impressive advancements in 3D point cloud filtering.
Point cloud filtering filters out noise by moving points in the noisy point cloud onto the underlying point set surfaces. 
In addition to removing noise, another important aspect is to preserve sharp geometric features in the point cloud.
The filtered point cloud  has applications in further geometry processing, rendering, computer-aided design, computer animation, etc. 

Among the existing point cloud filtering methods, LOP-relevant (locally optimal projection) techniques are robust to noise and can achieve good outcomes \cite{lipman2007parameterization,huang2009consolidation,preiner2014continuous}. However, they cannot well preserve sharp features, due to the lack of consideration of sharp feature information. RIMLS \cite{oztireli2009feature} and GPF \cite{lu2017gpf} introduced normal information to preserve sharp features better. Nevertheless, they rely heavily on the quality of normals, and poor normals would greatly impact the filtering outcomes. 
More recently, learning-based point cloud filtering techniques arose \cite{2020Pointfilter,Marie2020PointCleanNet,lu2020deep}. 
They can typically offer more automation. However, existing deep learning methods often require a large amount of training data. In other words, these methods will produce less decent results if they do not ``see'' enough samples in the training data. Also, they generally focus on local information and can hardly utilize non-local information. Some methods also depend on prior normal information to enable point cloud filtering \cite{2020Pointfilter,lu2020deep}. 
As shown in previous research \cite{lu2017gpf}, purely position-based methods like LOP cannot well preserve sharp features.

The above analysis motivates us to rethink feature-preserving point cloud filtering, from a non-learning and non-normal perspective. Similar local patterns frequently exist in many 3D point clouds, especially for complex models (e.g., relief). This paper attempts to model similar local patterns in a single 3D point cloud and aggregate them to generate feature-preserving point set surfaces without the dependence of normal information. In particular, we first define 3D local patches for each point 
(i.e. central point),
and then search the non-local similar patches of a local patch by comparing our designed similarity metric. 
In the second step, we aggregate the searched non-local patterns to achieve a filtered pattern. Provided that the local similar patches cannot be immediately aggregated due to the irregularity of points, we intuitively choose to merge the information of the central point position only in canonical space. The position of the central point will be calculated by averaging the central points of all similar patches in a canonical space and mapping it back to the original space. 
The pipeline can be iterated for a few times to achieve decent filtered outcomes, and we design two iteration schemes for use. We conduct extensive experiments on our method and various other methods, including position-based learning/non-learning methods and normal-based learning/non-learning methods. Results show that our method generally better preserves features than state-of-the-art position-based methods (i.e. without the use of normal information). It also achieves better or comparable results to normal-based methods.

The \textit{main technical contributions} of this paper are:
\begin{itemize}
    \item a non-learning non-local non-normal approach for feature-preserving point cloud filtering,
    \item a robust search algorithm for finding non-local similar  patches,
    \item an effective position update algorithm for fusing non-local similar information, and
    \item two iteration schemes for flexible use.
\end{itemize}

The \textit{main features} of this method are as follows.
\begin{itemize}
    \item User-friendliness. Our method is simple to implement and easy to use. \textit{We will release our source codes to the community.}
    \item Effectiveness. Our method is effective in smoothing the point set surfaces.
    \item Sharp feature preservation. Our method realizes the preservation of sharp features without taking account of normal information.
\end{itemize}
\section{Related Work}
\label{sec:relatedwork}
In this section, we briefly introduce previous works that are most related to our work. It includes point cloud filtering, non-local filtering methods, and robust principal component analysis.

\subsection{Point Cloud Filtering}
\label{sec:pointcloudsfilteringresearch}
The initial filtering of 3D point clouds is based on the idea of MLS (Moving Least Squares). 
\cite{levin1998approximation, levin2004mesh} introduced the conception of Moving Least Squares Method (MLS), which was used to smooth point cloud data. Later on, a series of variants of MLS have been proposed, such as \cite{alexa2001point,alexa2003computing,amenta2004defining,fleishman2005robust,oztireli2009feature}. Lange et al.\cite{lange2005anisotropic} developed a method for anisotropic fairing of a point sampled surface
using an anisotropic geometric mean curvature flow. 

Another family of point cloud filtering methods is motivated by Locally Optimal Projection (LOP) \cite{lipman2007parameterization}. LOP starts with downsampling the original point set into a second point set. It then projects this second point set onto the original point set by minimizing an objective function which consists of a $L_1$-median term (i.e. data term) and a repulsion term. 
Approaches based on LOP include weighted LOP (WLOP) \cite{huang2009consolidation}, kernel LOP (KLOP)  \cite{liao2013efficient}, continuous LOP (CLOP) \cite{preiner2014continuous} and anisotropic LOP (ALOP) \cite{huang2013edge}. WLOP encourages points to distribute more evenly, and CLOP makes the point set more compact at a faster speed by using Gaussian Mixture Model. ALOP preserves feature better with the aid of an anisotropic weighting function. 

To preserve sharp features, point cloud filtering methods often consider normal information. For example, Avron et al. \cite{avron2010} and Sun et al. \cite{sun2015denoising} designed $L_1$ and $L_0$ optimization considering normal information. 
GPF \cite{lu2017gpf}, which Gaussian Mixture Model inspired, has two terms and considers normal information to replace the point-to-point distance with point-to-plane distance. It preserves sharp edges well while filtering out noise. \cite{yifan2019differentiable} proposed a high-fidelity differentiable renderer based on point positions and normals, which could also be used in point cloud filtering. \cite{2020Lowrank} introduced a low-rank matrix approximation method to optimize the normal information before updating point positions. Liu et al. \cite{liu2020feature} proposed a feature-preserving method involving four stages (i.e. normal filtering, feature detection, multi-normal estimation, and point update). \cite{dinesh2020point} designed a local algorithm for 3D point cloud denoising. They designed a signal-dependent feature map Laplacian regularizer (SDFGLR) to improve the piecewise smoothness of surface normals to maintain sharp edges. 

More recently, a few deep learning methods have been proposed for denoising \cite{li2020dnf}, and showed promising performance in point cloud filtering \cite{roveri2018pointpronets,yu2018ec,Marie2020PointCleanNet,lu2020deep,2020Pointfilter,yu2018pu,yin2018p2p}. 
\cite{roveri2018pointpronets} transformed 3D points into 2D height maps, which are fed into CNNs for training. \cite{yu2018ec} first manually labeled sharp edges on 3D models and designed a network for consolidating point cloud data. 
PointCleanNet \cite{Marie2020PointCleanNet} proposed a two-stage method to separately remove outliers and  noise, without the consideration of normal information. DNP \cite{lu2020deep} was designed to first estimate the normals and then apply the point update algorithm in \cite{2020Lowrank}, thus achieving point cloud filtering. \cite{hermosilla2019total} proposed an unsupervised filtering method that only trains on noise data without ground truth. Due to the missing ground-truth information, they cannot well preserve sharp-edge features. \cite{pistilli2020learning} designed a graph-convolutional layers-based neural network to build hierarchies of features from similarity among the  high-dimensional feature representations of the point cloud.  \cite{luo2020differentiable} presented an autoencoder-like neural network. They obtain new sample points from the encoder step, infer the underlying manifold in decoder step, and then obtain a denoised point cloud by resampling on the reconstructed manifold.  \cite{hu2020feature} designed a feature graph learning algorithm that can also be used in 3D point cloud denoising.  \cite{2020Pointfilter} introduced a point-based deep learning framework for point cloud filtering, which elegantly preserves sharp edges with taking account of ground truth normals in the training phase only. 

\subsection{Non-local Filtering Methods}
\label{sec:nonlocalresearch}
Non-local methods have been widely used in filtering, such as image denoising \cite{zhu2017non,zhu2016non} and mesh denoising \cite{li2018non}. It has also been exploited in point cloud filtering. 
\cite{deschaud2010point} proposed a method using MLS surfaces as local descriptors to look for points that have similarities. The descriptor used by this method requires both point positions and normal information.   
\cite{digne2012similarity} proposed a denoising method based on local similarity, which smoothed the height vector field by comparing the neighborhood of one point with those of other points on the surface. It also relies highly on the quality of normals.
\cite{rosman2013patch} extended the Block-Matching 3D Denoising algorithm \cite{bm3d} to 3D point clouds, where the authors first obtained denoised surfaces via collaborative spectral shrinkage and then obtained denoised surfaces via collaborative spectral Wiener filtering. However, due to the long computation stages of the Iterative Closest Point (ICP) optimization and spectral decomposition calculation, a 15k point cloud requires several hours of denoising. 
\cite{digne2017sparse} defined a new description of the local patch named Local Probing Field (LPF). 
\cite{sarkar2018structured} divided the point cloud into patches and use the dictionary learning framework as a description of the patches. 
\cite{zeng20193d} extended a low-dimensional manifold model to represent patches of point cloud and designed a denoising method by seeking self-similar patches. To measure the distance between patches, they defined a reference plane obtained by normal information. Different from these methods which are inspired by MLS surfaces, height vector field, block matching, local probing field, etc, we attempt to design similarity through the RPCA decomposition and aggregate non-local similar information in the canonical space for position update (see Sections \ref{sec::overview}, \ref{sec::Similar_Patches_Finding} and \ref{sec::Position_Update}).

\subsection{Robust Principal Component Analysis}
\label{sec:rpcaresearch}
The Principle Component Analysis (PCA) is often regarded as a great tool in multi-dimensional data analysis. It is able to output a low-dimensional data representation from the raw high-dimensional data. On top of PCA, RPCA(Robust Principal Component Analysis) emerged \cite{wright2009robust} and is more robust than the original PCA analysis. It has been widely exploited in background modeling in video processing \cite{bouwmans2014robust,sobral2016lrslibrary}, 
image denoising \cite{lau2019restoration,chen2017denoising}, 
etc. In recent years, RPCA also has been used in 3D geometry processing. \cite{mattei2017point} designed an algorithm to filter the points by MRPCA (Moving Robust Principal Component Analysis), but they just estimated the points in a local way. \cite{chen2019multi} developed a multi-patch collaborative approach that defines a rotation-invariant height-map patch (HMP) for each point by robust Bi-PCA encoding bilaterally filtered normal information. Thus, they rely highly on the quality of normals and bilateral filtering.
In this work, we attempted to extend RPCA to design a similarity metric for locating non-local similar patches for feature-preserving point cloud filtering to get rid of the distribution caused by raw data noise. The main use of RPCA is to get non-local similar patches only from point position robustly.
\begin{figure*}[htbp]
\centering
\begin{minipage}[b]{0.95\linewidth}
{\label{}
\includegraphics[width=1\linewidth]{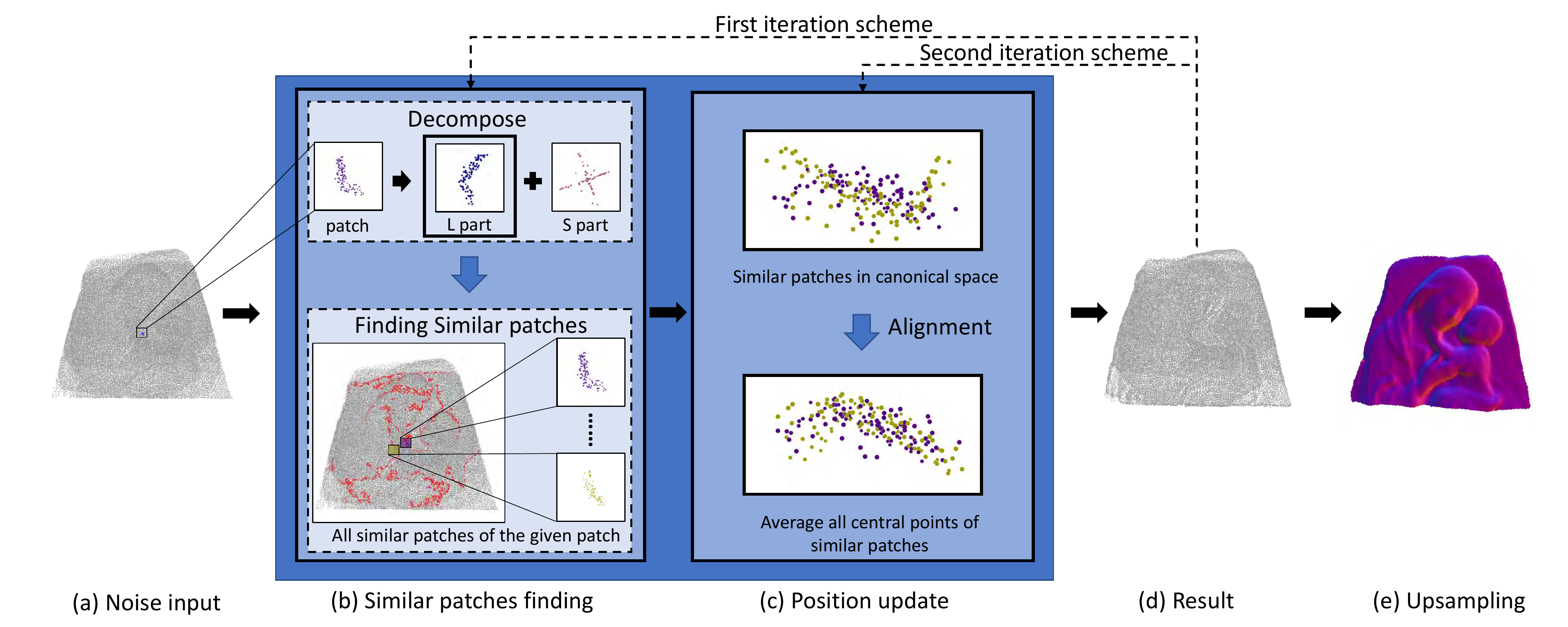}}
\end{minipage}
\caption{Overview of our approach. The approach consists of two steps: similar patches finding (b) and position update (c). The two iteration schemes respectively iterate (b)-(c) and (c) only for several times to achieve better filtering outcomes. (d) The filtering result. (e) The upsampling result. Our method is a purely position based method. }
\label{fig:overview}
\end{figure*}

\section{Overview}
\label{sec::overview}
The method we proposed includes two steps: similar patches finding and position update. We also discuss two iteration schemes in order to achieve better filtering results (Section \ref{sec::iterative_strategy}). Figure \ref{fig:overview} shows an overview of the proposed method.

\begin{itemize}
    \item \textbf{Similar patches finding} (Section \ref{sec::Similar_Patches_Finding}).  We find similar patches of a given local patch. Note that each point (i.e. ``central point'') and its neighbors construct a local patch. Specifically, we adopt RPCA to decompose the given local patch and obtain its low rank part which will be decomposed using SVD (Singular Value Decomposition). The three singular values define a vector for this given patch. We also perform the same procedure for all other local patches in the point cloud. We intuitively define the Euclidean distance between the pair of singular value vectors (from the given patch and another patch) as the similarity metric. We simply regard them as similar patches if the distance is lower than a threshold.
    
    \item \textbf{Position update} (Section \ref{sec::Position_Update}). We align the searched similar patches into a canonical coordinate space by defining a mapping function, and update the central point of the given patch by averaging the positions of all central points in different aligned patches, and finally re-map it to the original coordinate space with the inverse mapping. In this way, sharp features can be preserved by aggregating non-local similar information.
\end{itemize}

\section{Similar Patches Finding}
\label{sec::Similar_Patches_Finding}

In this section, we first give the definition of local patches in a point set and then formulate our non-local similarity problem by using RPCA. We utilize ALM (Augmented Lagrange Multiplier) to solve the convex optimization problem of RPCA \cite{wright2009robust}. We finally explain how to measure the similarity of two local patches.

Given a noisy point cloud $\mathbb{P}={\{p_i\}}_{i=1}^N \subset{R^3}$, we assume a point $p_i$ ($3\times1$ vector) and its neighboring points define a local patch $M$ ($3\times K$ matrix). The set $M$ is made up of the $K$ nearest neighbors of the given point $p_i$ (including this point). As a result, this point set has $N$ local patches in total. 

\textbf{RPCA formulation.} 
The key idea is to analyze the local patch using RPCA. To enable this, we firstly center the patch by subtracting the mean of all positions of the patch and then decompose a local patch $M$ into two parts $L$ and $S$, where $L$ is the low-rank part representing significant information and $S$ is the sparse part indicating perturbation information. Since there is no prior information on the low-rank part and the sparse part, it seems impossible to decompose the known matrix $M$ straightforwardly. 
According to the basic PCP (Principal Component Pursuit) assumption \cite{2009Rank}, it is feasible to get the constrained optimization as:
\begin{equation}\label{eq:principal_component_pursult}
\begin{aligned}
\min{\|L\|}_*+\lambda{\|S\|}_1,~ \text{subject to} \ \ {M}=L+S,
\end{aligned}
\end{equation}
where ${\|\cdot\|}_*$ is the nuclear norm (the sum of all singular values of a matrix), ${\|\cdot\|}_1$ is the sum of absolute values of all elements in a matrix, and $\lambda$ is a weighting parameter. 

\textbf{Optimization.}
It is obvious that Eq. \eqref{eq:principal_component_pursult} is a convex optimization problem. 
Among the methods to solve convex optimization of  RPCA \cite{ma2018efficient}, we choose the Alternating Lagrange Multiplier (ALM) based algorithm introduced by \cite{Lin2010The} to solve the problem since ALM always achieves high accuracy in fewer iterations. According to Emmanuel et al. \cite{Emmanuel2009Exact}, it can also work stably in a variety of problem settings without adjusting parameters. 

From Algorithm \ref{alg:Principal_Component_Pursuit_BY_ALM}, we can decompose the given patch $M$ into a low-rank part $L$ and a sparse part $S$ \cite{Lin2010The,yuan2009sparse}.
${\mathcal{D}_\tau}()$ and ${\mathcal{S}_\tau}()$ respectively denote the shrinkage operator and the singular value thresholding, operator \cite{2009Robust_Book}. Here, we choose the low rank part $L$ as the main representation of $M$, since $L$ is supposed to be the significant information and $S$ is the perturbation information  as revealed by RPCA. Thus we use $L$ for the following calculation.

\begin{algorithm}[h]  
  \caption{Matrix Decomposition ($MD$)
  }  
  \label{alg:Principal_Component_Pursuit_BY_ALM}  
  \begin{algorithmic}[1]
    \Require matrix $M$
    \State $\textbf{Initialize:}  S=Y=0,\mu >0$
    \While{not converged}
      \State  $L'={\mathcal{D}_{\frac{1}{\mu}}}(M-S+\mu^{-1}Y)$  
      \State $S'=\mathcal{S}_\frac{\lambda}{\mu}(M-L'+\mu^{-1}Y)$ 
      \State $Y'=Y+\mu(M-L'-S')$
      \State $L=L'$, $S=S'$, $Y=Y'$
    \EndWhile
    \Ensure $L$, $S$
  \end{algorithmic}  
\end{algorithm} 

\textbf{Similarity.} We calculate the singular values of $L$ and take them as the representation of the given local patch $M$. In other words, each local patch $M$ can be represented as a vector $v$ ($1 \times 3$ vector) composed of three singular values of $L$. Given two local patches $M$ and $M'$, we first calculate the representation vector $v$ obtained by $M$ and vector $v'$ obtained by $M'$. We then calculate the Euclidean distance of these two vectors as the similarity (or difference) between the two local patches. 
\begin{equation}\label{eq:similarity}
\begin{aligned}
d = ||v-v'||,
\end{aligned}
\end{equation}
where $d$ is the similarity between the two patches $M$ and $M'$. 
Further, we define a threshold $\theta$ to judge whether $M'$ is similar to $M$, shown in Eq. \eqref{eq:similarornot}. The two patches are similar if $d<\theta$, and dissimilar otherwise. In this way, we can achieve each local patch's similar patches in the point cloud.
\begin{align}
\label{eq:similarornot}
   \begin{array}{l}
     sim = \left\{\begin{array}{l}1, \text{ if } d<\theta \\\text 0, \text{ else}\end{array}\right . 
\end{array} 
\end{align}

\begin{figure}[htb]
\centering
\begin{minipage}[b]{0.4\linewidth}
\subfigure[]{\label{}\includegraphics[width=1\linewidth]{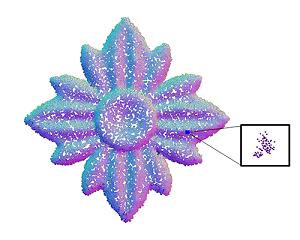}}
\end{minipage}
\begin{minipage}[b]{0.4\linewidth}
\subfigure[]{\label{}\includegraphics[width=1\linewidth]{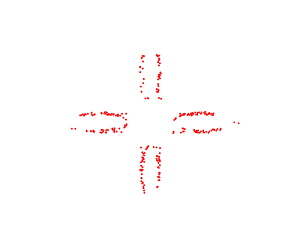}}
\end{minipage}
\\
\begin{minipage}[b]{0.4\linewidth}
\subfigure[]{\label{}\includegraphics[width=1\linewidth]{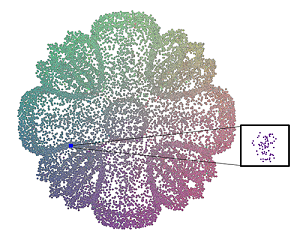}}
\end{minipage}
\begin{minipage}[b]{0.4\linewidth}
\subfigure[]{\label{}\includegraphics[width=1\linewidth]{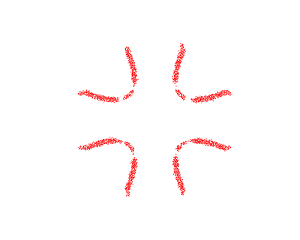}}
\end{minipage}
\caption{Similar patches of two given patches. (a) and (c): different central points (blue points) define different patches (blown-up windows). (b) and (d): the central points of the similar patches searched by our method (a red point denotes a local patch) according to the given patch. }
\label{fig:find_similar_patch}
\end{figure}

Algorithm \ref{alg:similar_patch_finding} finds all similar patches for the given local patch $M$. We assume that the shapes do have similar patterns and employ RPCA (more robust than PCA) for the design of the similarity metric. Under suitable radius and threshold conditions, we are able to extract similar patches. To illustrate the effectiveness of our method, we use our method to compute similar patches of given local patches on two different models in Figure \ref{fig:find_similar_patch}, and we can see that our similarity metric works very well. 

\begin{algorithm}[h]  
  \caption{Similar Patches Finding }  
  \label{alg:similar_patch_finding}  
  \begin{algorithmic}[1]
    \Require local patch $M$, threshold $\theta$
    \State \textbf{Initialize:} similar patches set $T$
    \State$L,S=MD(M)$
    \State$v=SVD(L)$
    \For{each patch $M'$ \textbf{in} point cloud}
    \State$L',S'=MD(M')$
    \State$v'=SVD(L')$
    \State$d=\|v-v'\|$
        \If{$d < \theta$}
            \State add $M'$ to $T$
        \EndIf
    \EndFor
    \Ensure the similar patches set $T$ to patch $M$
  \end{algorithmic}  
\end{algorithm}

\section{Position Update}
\label{sec::Position_Update}
After finding similar patches, we need to further update point positions in order to remove noise. We design an effective non-local mean algorithm to update point positions.

Provided the similar patches set $T$ in Section \ref{sec::Similar_Patches_Finding}, the positions and the poses of them are various in the original space, which makes it impossible to ``average'' similar patches like image processing. Given this, we attempt to average the central points of similar patches, which simplifies the problem and makes it feasible to update point positions. It requires aligning similar patches into a canonical space before averaging.

Suppose $D = M M^T \in R^{3 \times 3}$ represents the covariance matrix of the given matrix $M$.  $D$ can be decomposed into three eigenvectors (as column vectors) which together define a $3\times3$ matrix $F$. We use $F$ as the mapping function, and multiply it with the given patch $M$ to obtain a mapped local patch $M_f$ in the ``eigen'' space (i.e. canonical space).
\begin{equation}\label{eq:mappingtoCanonical}
    \begin{aligned}
        {M_f} = {F} {M}
    \end{aligned}
\end{equation}
We use the above equation to map the similar patches in $T$ to get the mapped similar patches set $T_f$.

\begin{figure}[htb]
\centering
\begin{minipage}[b]{0.32\linewidth}
\subfigure[]{\label{}\includegraphics[width=1\linewidth]{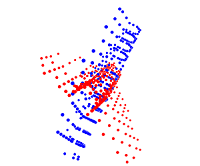}}
\end{minipage}
\begin{minipage}[b]{0.32\linewidth}
\subfigure[]{\label{}\includegraphics[width=1\linewidth]{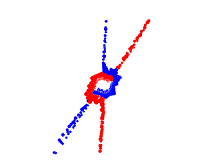}}
\end{minipage}
\begin{minipage}[b]{0.32\linewidth}
\subfigure[]{\label{}\includegraphics[width=1\linewidth]{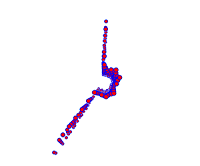}}
\end{minipage} 
\caption{Aligning patches in canonical space. (a) Two similar patches (blue points, red points) in the original space. (b) They are mapped to a canonical space (another view). (c) After regularization using Eq. \eqref{eq:match_patch_by_eight}. }
\label{fig:align_in_canonical}
\end{figure}

In theory, these similar patches should be aligned in this canonical space. However, it is impractical to fully align each local patch of points like image patches, due to the irregularity of points, noise, and directions of the eigen matrix mapping (positive and negative). But we can try to mitigate the direction  issue. Figure \ref{fig:align_in_canonical} gives an example. 
Because the mapping direction is random on the positive and negative sides of the coordinate axes, we generate a total of eight patches $\{O_i| i=1,2,...,8\}$ by flipping the patch $O$ ($O \in T_f$) around three coordinate axes in the canonical space (including the original patch). This is because 3 axes form 8 quadrants. We target to find a patch in $\{O_i\}$ which is most aligned to the given patch $M_f$. Firstly, we divide each patch $O_i$ into 8 sub-patches $\{O_i^j|i,j=1,2,..,8\}$ according to the axes. To reduce the runtime, we simply use PCA instead of RPCA to calculate the principal components of each sub-patch. For two identical patches which are exactly the same and aligned, the eigenvector direction of the sub-patches in each quadrant should be consistent or opposite. 
Considering the 8 sub-patch eigenvectors at the same time can effectively alleviate the randomness of positive and negative signs of patch $O_i$. Therefore, we define the offset ($\sigma_i$) between patch $O_i$ and patch $M_f$ as
\begin{equation}\label{eq:match_patch_by_eight}
\begin{aligned}
\sigma_i = \sum_{j=1}^8{\min{(\|v_{O_i^j} - v_j\|_2,\|v_{O_i^j} + v_j\|_2})},
\end{aligned}
\end{equation}
where $v_{O^i_j}$ represents the eigen vector of the sub-patch in the $j$-th quadrant spilt from patch $O_i$. $v_j$ represents the eigen vector of the sub-patch in the $j$-th quadrant spilt from patch $M_f$. We select the patch $O_i$ with the minimum $\sigma_i$ as the most similar patch to $M_f$, and update patch $O$ with patch $O_i$.

With the above processing, we obtain the regularized similar patches set $T_r$. In the canonical space, we simply update the new position of the central point of patch $M_f$ by calculating the mean coordinates of all central points of patches in $T_r$. 
\begin{equation}\label{}
\begin{aligned}
p_i^f = \frac{\sum_jp_{i,j}^f}{|T_r|},
\end{aligned}
\end{equation}
where $\{p_{i,j}^f\}$ denotes the set of positions of all centred points in the similar patches set $T_r$ in the canonical space, and $p_i^f$ indicates the mean value. $|T_r|$ is the number of similar patches in the set. 

At last, we re-map the new position from the canonical space to the original space by multiplying the inverse mapping matrix.
\begin{equation}\label{}
\begin{aligned}
p_i^{new} = F_{inv} p_i^f,
\end{aligned}
\end{equation}
where $F_{inv}$ denotes the inverse matrix of $F$ and $p_i^{new}$ represents the new position in the original space. 

\section{Iteration Schemes}
\label{sec::iterative_strategy}
We often obtain better filtering results by performing a few iterations of the above two steps. We design two iteration schemes for our method. One is to re-find similar patches of each given patch in each iteration, and the other is simply using the searched similar patches in the first iteration in follow-up iterations. It is obvious that the second scheme reduces computation and increases speed. Since the first scheme regenerates similar patches in each iteration and the next iteration depends on the smoothed result in the current iteration, it would tend to eliminate noise more effectively than the second iteration scheme.

\begin{figure}[htb]
\centering
\begin{minipage}[b]{0.35\linewidth}
{\label{fig:iterative_strategy_syn1}\includegraphics[width=1\linewidth]{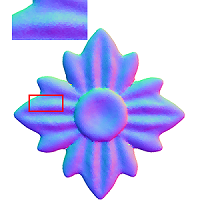}}
\end{minipage}
\begin{minipage}[b]{0.35\linewidth}
{\label{fig:iterative_strategy_syn2}\includegraphics[width=1\linewidth]{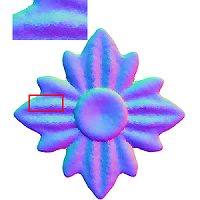}}
\end{minipage}
\\
\begin{minipage}[b]{0.35\linewidth}
{\label{fig:iterative_strategy_syn3}\includegraphics[width=1\linewidth]{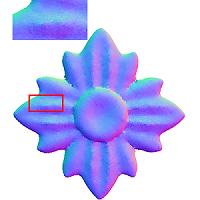}}
\end{minipage} 
\begin{minipage}[b]{0.35\linewidth}
{\label{fig:iterative_strategy_syn4}\includegraphics[width=1\linewidth]{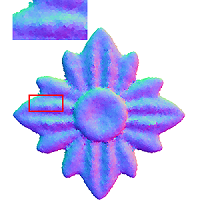}}
\end{minipage}
\\
\begin{minipage}[b]{0.35\linewidth}
\subfigure[First iteration scheme]
{\label{fig:iterative_strategy_raw1}\includegraphics[width=1\linewidth]{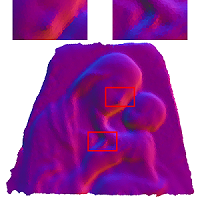}}
\end{minipage} 
\begin{minipage}[b]{0.35\linewidth}
\subfigure[Second iteration scheme]
{\label{fig:iterative_strategy_raw2}\includegraphics[width=1\linewidth]{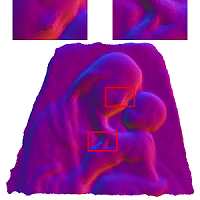}}
\end{minipage}
\caption{Comparisons for the two iteration schemes. First row: input is having 0.5\% noise. Second row: input is having 1.0\% noise. Third row: raw scan model.  }
\label{fig:iterative_strategy}
\end{figure}

We show the difference between the two iteration schemes in Figure \ref{fig:iterative_strategy}. We can see from the figure that the second scheme generates better results than the first scheme in preserving sharp features on point cloud with relatively small level of noise. The first scheme produces better results than the second scheme in terms of removing noise on point cloud with larger noise. In the raw scanned model, the second scheme is better than the first one, in terms of preserving sharp features. This is consistent with the first row (small synthetic noise) because this raw point scan is corrupted with small-scale noise.

In general, we will know the level of synthetic noise and then the automation of choice is straightforward. Regarding raw scanned noise, we have to manually judge that. Alternatively, it is plausible to assume the raw scanned noise is relatively low-level. To sum up, we conclude the following principles which will be used for experiments.
\begin{itemize}
    \item We use the first iteration scheme for point clouds with  
    relatively larger noise.
    \item Otherwise we employ the second iteration scheme.
\end{itemize}

\section{Experimental Results}
In this section, we first describe the experimental setup and then compare our method with non-deep-learning methods and deep learning methods (visual comparisons and quantitative comparisons). We next conduct ablation studies of the proposed method and finally discuss its limitations and future work.

\subsection{Experimental Setup}
\label{sec::parameter_setting}
\textbf{Parameter setting.}
Note that we use the fixed parameters for RPCA as suggested by the original work \cite{2009Robust_Book}. Thus, the involved parameters of our method are: patch size $K$, the similarity threshold $\theta$, and the number of iterations for the method. We use KNN to search the local neighbors for each point, and we set a larger value for a greater number of points in the point cloud ($K\in[50,150]$).
The threshold $\theta$ is used to control the number of similar patches ($\theta \in[0.01,0.5]$). A larger $\theta$ will generally enable more similar patches and a smaller $\theta$ will induce less similar patches. Besides, we propose two iteration schemes (see Section \ref{sec::iterative_strategy}), and set 1-3 iterations to our method.

\begin{figure*}[htbp]
\centering
\begin{minipage}[b]{0.13\linewidth}
{\label{}\includegraphics[width=1\linewidth]{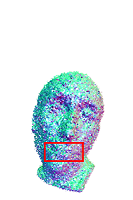}}
\end{minipage}
\begin{minipage}[b]{0.13\linewidth}
{\label{}\includegraphics[width=1\linewidth]{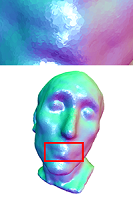}}
\end{minipage}
\begin{minipage}[b]{0.13\linewidth}
{\label{}\includegraphics[width=1\linewidth]{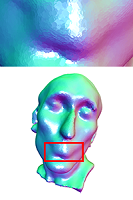}}
\end{minipage}
\begin{minipage}[b]{0.13\linewidth}
{\label{}\includegraphics[width=1\linewidth]{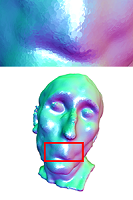}}
\end{minipage}
\begin{minipage}[b]{0.13\linewidth}
{\label{}\includegraphics[width=1\linewidth]{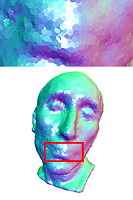}}
\end{minipage}
\begin{minipage}[b]{0.13\linewidth}
{\label{}\includegraphics[width=1\linewidth]{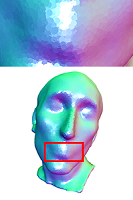}}
\end{minipage} 
\begin{minipage}[b]{0.13\linewidth}
{\label{}\includegraphics[width=1\linewidth]{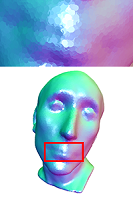}}
\end{minipage} 
\\
\vspace{0.3cm}
\begin{minipage}[b]{0.13\linewidth}
{\label{}\includegraphics[width=1\linewidth]{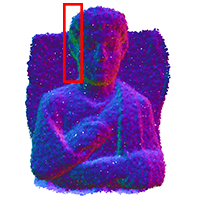}}
\end{minipage}
\begin{minipage}[b]{0.13\linewidth}
{\label{}\includegraphics[width=1\linewidth]{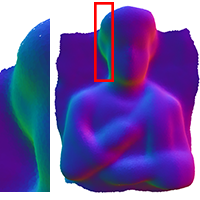}}
\end{minipage}
\begin{minipage}[b]{0.13\linewidth}
{\label{}\includegraphics[width=1\linewidth]{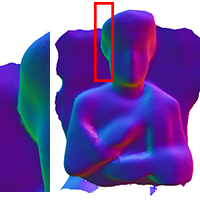}}
\end{minipage}
\begin{minipage}[b]{0.13\linewidth}
{\label{}\includegraphics[width=1\linewidth]{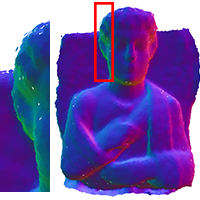}}
\end{minipage}
\begin{minipage}[b]{0.13\linewidth}
{\label{}\includegraphics[width=1\linewidth]{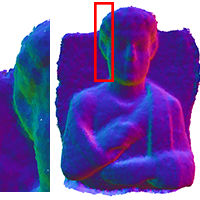}}
\end{minipage}
\begin{minipage}[b]{0.13\linewidth}
{\label{}\includegraphics[width=1\linewidth]{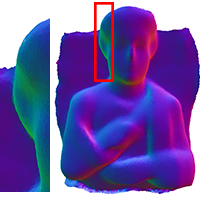}}
\end{minipage} 
\begin{minipage}[b]{0.13\linewidth}
{\label{}\includegraphics[width=1\linewidth]{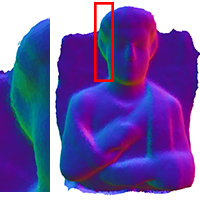}}
\end{minipage} 
\\
\vspace{0.3cm}
\begin{minipage}[b]{0.13\linewidth}
{\label{}\includegraphics[width=1\linewidth]{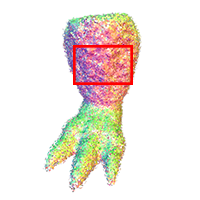}}
\end{minipage}
\begin{minipage}[b]{0.13\linewidth}
{\label{}\includegraphics[width=1\linewidth]{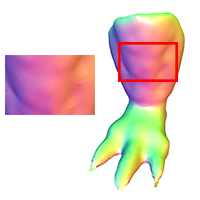}}
\end{minipage}
\begin{minipage}[b]{0.13\linewidth}
{\label{}\includegraphics[width=1\linewidth]{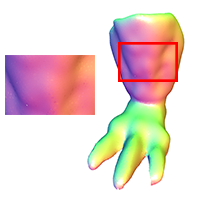}}
\end{minipage}
\begin{minipage}[b]{0.13\linewidth}
{\label{}\includegraphics[width=1\linewidth]{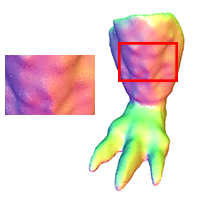}}
\end{minipage}
\begin{minipage}[b]{0.13\linewidth}
{\label{}\includegraphics[width=1\linewidth]{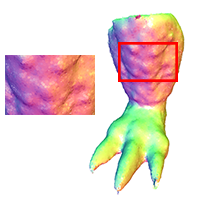}}
\end{minipage}
\begin{minipage}[b]{0.13\linewidth}
{\label{}\includegraphics[width=1\linewidth]{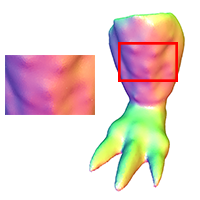}}
\end{minipage} 
\begin{minipage}[b]{0.13\linewidth}
{\label{}\includegraphics[width=1\linewidth]{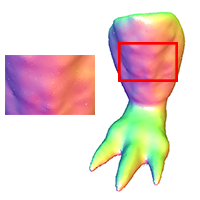}}
\end{minipage} 
\\
\vspace{0.3cm}
\begin{minipage}[b]{0.13\linewidth}
{\label{}\includegraphics[width=1\linewidth]{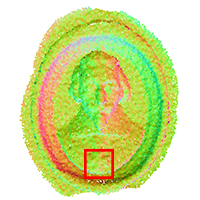}}
\end{minipage}
\begin{minipage}[b]{0.13\linewidth}
{\label{}\includegraphics[width=1\linewidth]{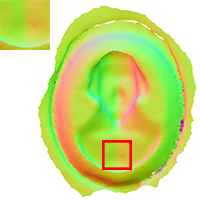}}
\end{minipage}
\begin{minipage}[b]{0.13\linewidth}
{\label{}\includegraphics[width=1\linewidth]{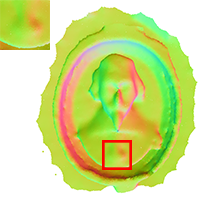}}
\end{minipage}
\begin{minipage}[b]{0.13\linewidth}
{\label{}\includegraphics[width=1\linewidth]{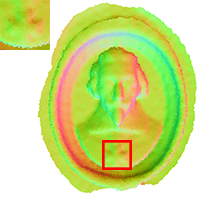}}
\end{minipage}
\begin{minipage}[b]{0.13\linewidth}
{\label{}\includegraphics[width=1\linewidth]{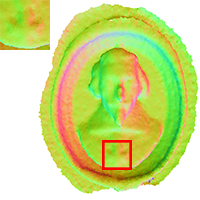}}
\end{minipage}
\begin{minipage}[b]{0.13\linewidth}
{\label{}\includegraphics[width=1\linewidth]{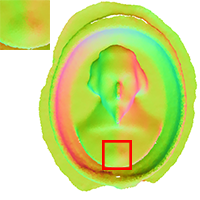}}
\end{minipage} 
\begin{minipage}[b]{0.13\linewidth}
{\label{}\includegraphics[width=1\linewidth]{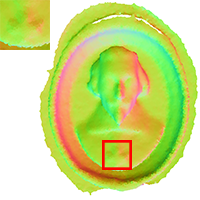}}
\end{minipage} 
\\
\vspace{0.3cm}
\begin{minipage}[b]{0.13\linewidth}
{\label{}\includegraphics[width=1\linewidth]{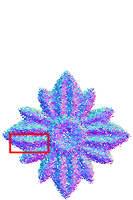}}
\end{minipage}
\begin{minipage}[b]{0.13\linewidth}
{\label{}\includegraphics[width=1\linewidth]{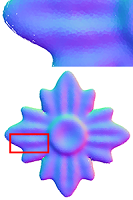}}
\end{minipage}
\begin{minipage}[b]{0.13\linewidth}
{\label{}\includegraphics[width=1\linewidth]{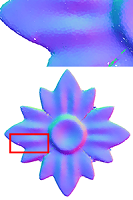}}
\end{minipage}
\begin{minipage}[b]{0.13\linewidth}
{\label{}\includegraphics[width=1\linewidth]{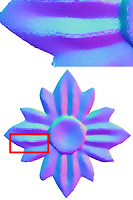}}
\end{minipage}
\begin{minipage}[b]{0.13\linewidth}
{\label{}\includegraphics[width=1\linewidth]{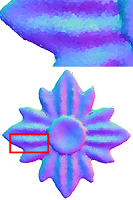}}
\end{minipage}
\begin{minipage}[b]{0.13\linewidth}
{\label{}\includegraphics[width=1\linewidth]{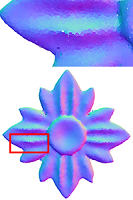}}
\end{minipage} 
\begin{minipage}[b]{0.13\linewidth}
{\label{}\includegraphics[width=1\linewidth]{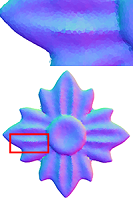}}
\end{minipage} 
\\
\vspace{0.3cm}
\begin{minipage}[b]{0.13\linewidth}
{\label{}\includegraphics[width=1\linewidth]{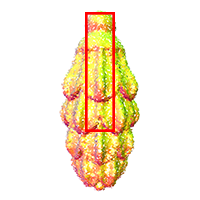}}
\end{minipage}
\begin{minipage}[b]{0.13\linewidth}
{\label{}\includegraphics[width=1\linewidth]{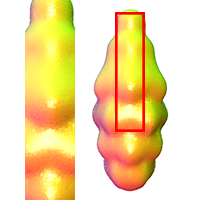}}
\end{minipage}
\begin{minipage}[b]{0.13\linewidth}
{\label{}\includegraphics[width=1\linewidth]{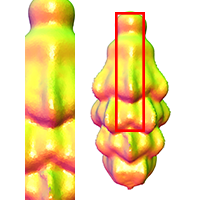}}
\end{minipage}
\begin{minipage}[b]{0.13\linewidth}
{\label{}\includegraphics[width=1\linewidth]{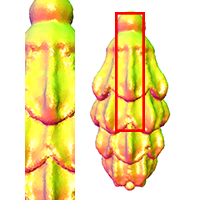}}
\end{minipage}
\begin{minipage}[b]{0.13\linewidth}
{\label{}\includegraphics[width=1\linewidth]{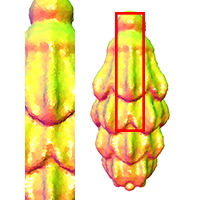}}
\end{minipage}
\begin{minipage}[b]{0.13\linewidth}
{\label{}\includegraphics[width=1\linewidth]{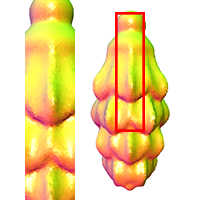}}
\end{minipage} 
\begin{minipage}[b]{0.13\linewidth}
{\label{}\includegraphics[width=1\linewidth]{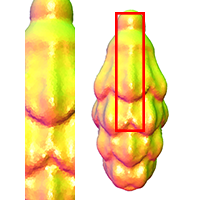}}
\end{minipage} 
\\
\vspace{0.3cm}
\begin{minipage}[b]{0.13\linewidth}
{\label{}\includegraphics[width=1\linewidth]{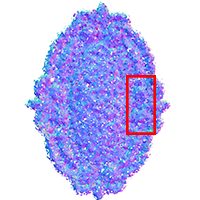}}
\centerline{Noisy input}
\end{minipage}
\begin{minipage}[b]{0.13\linewidth}
{\label{}\includegraphics[width=1\linewidth]{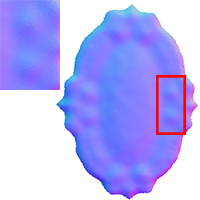}}
\centerline{CLOP \cite{preiner2014continuous}}
\end{minipage}
\begin{minipage}[b]{0.13\linewidth}
{\label{}\includegraphics[width=1\linewidth]{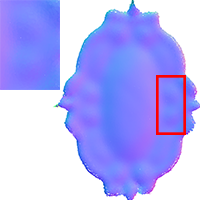}}
\centerline{GPF \cite{lu2017gpf}}
\end{minipage}
\begin{minipage}[b]{0.13\linewidth}
{\label{}\includegraphics[width=1\linewidth]{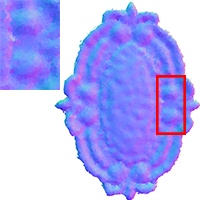}}
\centerline{RIMLS \cite{oztireli2009feature}}
\end{minipage}
\begin{minipage}[b]{0.13\linewidth}
{\label{}\includegraphics[width=1\linewidth]{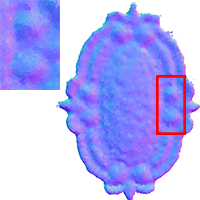}}
\centerline{PointCleanNet \cite{Marie2020PointCleanNet}}
\end{minipage}
\begin{minipage}[b]{0.13\linewidth}
{\label{}\includegraphics[width=1\linewidth]{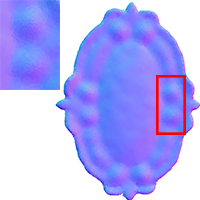}}
\centerline{PointFilter \cite{2020Pointfilter}}
\end{minipage} 
\begin{minipage}[b]{0.13\linewidth}
{\label{}\includegraphics[width=1\linewidth]{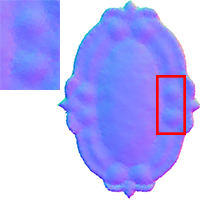}}
\centerline{Ours}
\end{minipage} 
\\
\caption{ Filtering results on 7 synthetic point clouds. From top to bottom: Nicolo (1.0\% noise), Funeral man (0.5\% noise), Leg (1.0\% noise), Richelieu (0.5\% noise), Flower (1.0\% noise), Fruit (0.5\% noise) and Mirror (1.0\% noise). }
\label{fig:synthetic}
\end{figure*}

\begin{figure*}[htbp]
\centering
\begin{minipage}[b]{0.13\linewidth}
{\label{}\includegraphics[width=1\linewidth]{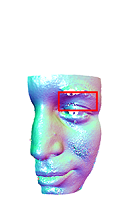}}
\end{minipage}
\begin{minipage}[b]{0.13\linewidth}
{\label{}\includegraphics[width=1\linewidth]{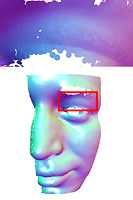}}
\end{minipage}
\begin{minipage}[b]{0.13\linewidth}
{\label{}\includegraphics[width=1\linewidth]{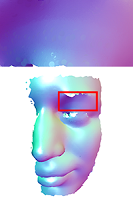}}
\end{minipage}
\begin{minipage}[b]{0.13\linewidth}
{\label{}\includegraphics[width=1\linewidth]{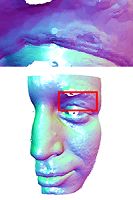}}
\end{minipage}
\begin{minipage}[b]{0.13\linewidth}
{\label{}\includegraphics[width=1\linewidth]{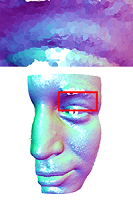}}
\end{minipage}
\begin{minipage}[b]{0.13\linewidth}
{\label{}\includegraphics[width=1\linewidth]{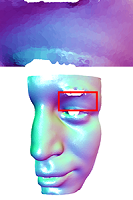}}
\end{minipage} 
\begin{minipage}[b]{0.13\linewidth}
{\label{}\includegraphics[width=1\linewidth]{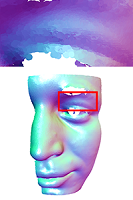}}
\end{minipage} 
\\
\begin{minipage}[b]{0.13\linewidth}
{\label{}\includegraphics[width=1\linewidth]{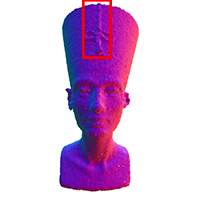}}
\end{minipage}
\begin{minipage}[b]{0.13\linewidth}
{\label{}\includegraphics[width=1\linewidth]{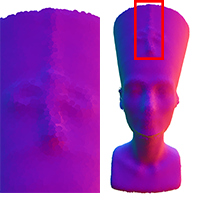}}
\end{minipage}
\begin{minipage}[b]{0.13\linewidth}
{\label{}\includegraphics[width=1\linewidth]{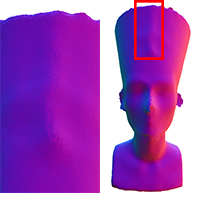}}
\end{minipage}
\begin{minipage}[b]{0.13\linewidth}
{\label{}\includegraphics[width=1\linewidth]{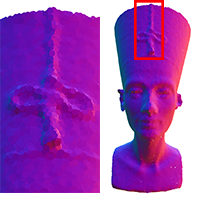}}
\end{minipage}
\begin{minipage}[b]{0.13\linewidth}
{\label{}\includegraphics[width=1\linewidth]{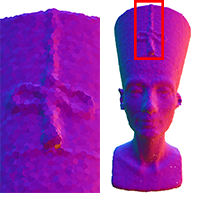}}
\end{minipage}
\begin{minipage}[b]{0.13\linewidth}
{\label{}\includegraphics[width=1\linewidth]{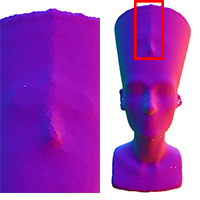}}
\end{minipage} 
\begin{minipage}[b]{0.13\linewidth}
{\label{}\includegraphics[width=1\linewidth]{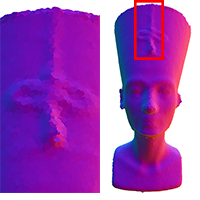}}
\end{minipage} 
\\
\begin{minipage}[b]{0.13\linewidth}
{\label{}\includegraphics[width=1\linewidth]{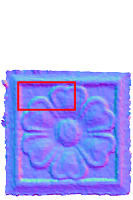}}
\end{minipage}
\begin{minipage}[b]{0.13\linewidth}
{\label{}\includegraphics[width=1\linewidth]{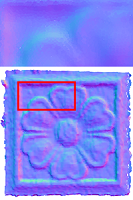}}
\end{minipage}
\begin{minipage}[b]{0.13\linewidth}
{\label{}\includegraphics[width=1\linewidth]{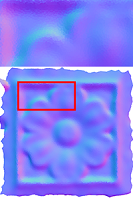}}
\end{minipage}
\begin{minipage}[b]{0.13\linewidth}
{\label{}\includegraphics[width=1\linewidth]{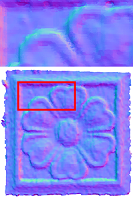}}
\end{minipage}
\begin{minipage}[b]{0.13\linewidth}
{\label{}\includegraphics[width=1\linewidth]{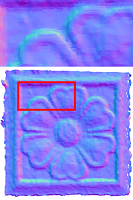}}
\end{minipage}
\begin{minipage}[b]{0.13\linewidth}
{\label{}\includegraphics[width=1\linewidth]{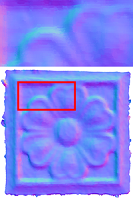}}
\end{minipage} 
\begin{minipage}[b]{0.13\linewidth}
{\label{}\includegraphics[width=1\linewidth]{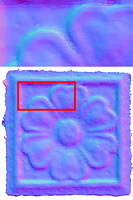}}
\end{minipage} 
\\
\begin{minipage}[c]{0.13\linewidth}
{\label{}\includegraphics[width=1\linewidth]{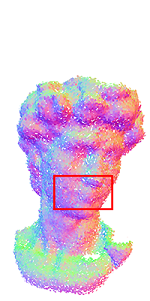}}
\end{minipage}
\begin{minipage}[c]{0.13\linewidth}
{\label{}\includegraphics[width=1\linewidth]{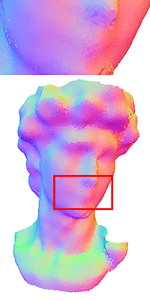}}
\end{minipage}
\begin{minipage}[c]{0.13\linewidth}
{\label{}\includegraphics[width=1\linewidth]{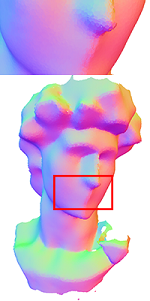}}
\end{minipage}
\begin{minipage}[c]{0.13\linewidth}
{\label{}\includegraphics[width=1\linewidth]{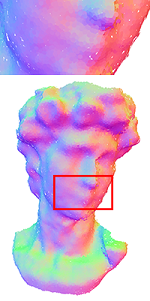}}
\end{minipage}
\begin{minipage}[c]{0.13\linewidth}
{\label{}\includegraphics[width=1\linewidth]{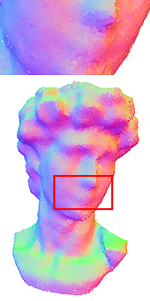}}
\end{minipage}
\begin{minipage}[c]{0.13\linewidth}
{\label{}\includegraphics[width=1\linewidth]{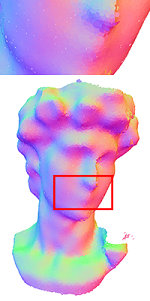}}
\end{minipage} 
\begin{minipage}[c]{0.13\linewidth}
{\label{}\includegraphics[width=1\linewidth]{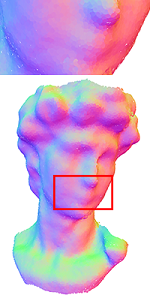}}
\end{minipage} 
\\
\begin{minipage}[c]{0.13\linewidth}
{\label{}\includegraphics[width=1\linewidth]{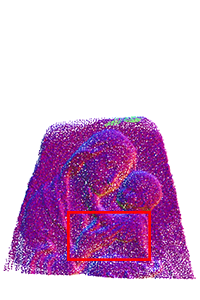}}
\end{minipage}
\begin{minipage}[c]{0.13\linewidth}
{\label{}\includegraphics[width=1\linewidth]{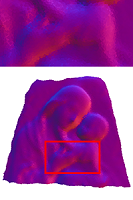}}
\end{minipage}
\begin{minipage}[c]{0.13\linewidth}
{\label{}\includegraphics[width=1\linewidth]{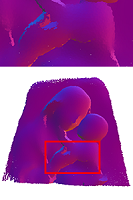}}
\end{minipage}
\begin{minipage}[c]{0.13\linewidth}
{\label{}\includegraphics[width=1\linewidth]{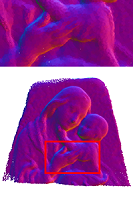}}
\end{minipage}
\begin{minipage}[c]{0.13\linewidth}
{\label{}\includegraphics[width=1\linewidth]{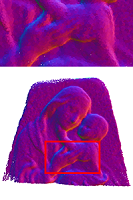}}
\end{minipage}
\begin{minipage}[c]{0.13\linewidth}
{\label{}\includegraphics[width=1\linewidth]{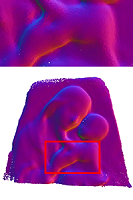}}
\end{minipage} 
\begin{minipage}[c]{0.13\linewidth}
{\label{}\includegraphics[width=1\linewidth]{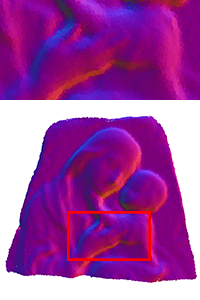}}
\end{minipage} 
\\
\begin{minipage}[c]{0.13\linewidth}
{\label{}\includegraphics[width=1\linewidth]{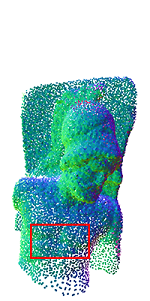}}
\centerline{Noisy input}
\end{minipage}
\begin{minipage}[c]{0.13\linewidth}
{\label{}\includegraphics[width=1\linewidth]{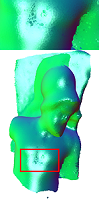}}
\centerline{CLOP \cite{preiner2014continuous}}
\end{minipage}
\begin{minipage}[c]{0.13\linewidth}
{\label{}\includegraphics[width=1\linewidth]{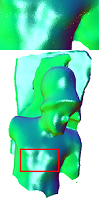}}
\centerline{GPF \cite{lu2017gpf}}
\end{minipage}
\begin{minipage}[c]{0.13\linewidth}
{\label{}\includegraphics[width=1\linewidth]{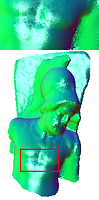}}
\centerline{RIMLS \cite{oztireli2009feature}}
\end{minipage}
\begin{minipage}[c]{0.13\linewidth}
{\label{}\includegraphics[width=1\linewidth]{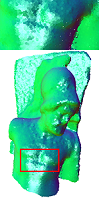}}
\centerline{PointCleanNet \cite{Marie2020PointCleanNet}}
\end{minipage}
\begin{minipage}[c]{0.13\linewidth}
{\label{}\includegraphics[width=1\linewidth]{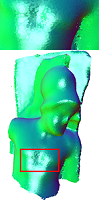}}
\centerline{PointFilter \cite{2020Pointfilter}}
\end{minipage} 
\begin{minipage}[c]{0.13\linewidth}
{\label{}\includegraphics[width=1\linewidth]{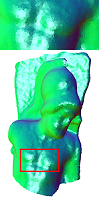}}
\centerline{Ours}
\end{minipage} 
\\
\caption{ Filtering results on 6 raw scan models. From top to bottom: Face, Nefertiti, Rawrelief, David, Relief, and Soldier.  }
\label{fig:raw}
\end{figure*}

\begin{figure*}[htb]
\centering
\begin{minipage}[b]{0.15\linewidth}
{\label{}\includegraphics[width=1\linewidth]{paper/figures/virtual_scan/david/noisy.png}}
\end{minipage}
\begin{minipage}[b]{0.15\linewidth}
{\label{}\includegraphics[width=1\linewidth]{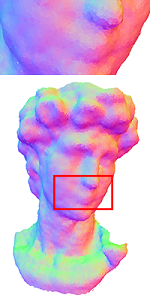}}
\end{minipage}
\begin{minipage}[b]{0.15\linewidth}
{\label{}\includegraphics[width=1\linewidth]{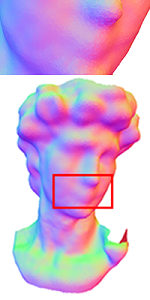}}
\end{minipage}
\begin{minipage}[b]{0.15\linewidth}
{\label{}\includegraphics[width=1\linewidth]{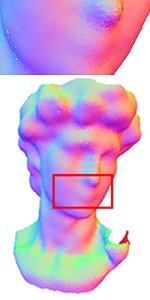}}
\end{minipage}
\begin{minipage}[b]{0.15\linewidth}
{\label{}\includegraphics[width=1\linewidth]{paper/figures/virtual_scan/david/ours.png}}
\end{minipage} 
\\
\begin{minipage}[b]{0.15\linewidth}
{\label{}\includegraphics[width=1\linewidth]{paper/figures/virtual_scan/relief_c/noisy.png}}
\end{minipage}
\begin{minipage}[b]{0.15\linewidth}
{\label{}\includegraphics[width=1\linewidth]{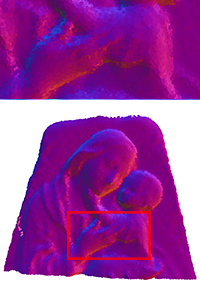}}
\end{minipage}
\begin{minipage}[b]{0.15\linewidth}
{\label{}\includegraphics[width=1\linewidth]{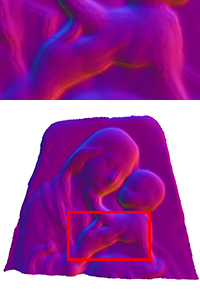}}
\end{minipage}
\begin{minipage}[b]{0.15\linewidth}
{\label{}\includegraphics[width=1\linewidth]{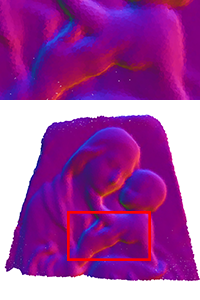}}
\end{minipage}
\begin{minipage}[b]{0.15\linewidth}
{\label{}\includegraphics[width=1\linewidth]{paper/figures/virtual_scan/relief_c/ours.png}}
\end{minipage} 
\\
\begin{minipage}[b]{0.15\linewidth}
{\label{}\includegraphics[width=1\linewidth]{paper/figures/virtual_scan/soldier/noisy.png}}
\centerline{Noisy input}
\end{minipage}
\begin{minipage}[b]{0.15\linewidth}
{\label{}\includegraphics[width=1\linewidth]{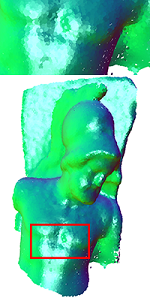}}
\centerline{DNP \cite{lu2020deep}}
\end{minipage}
\begin{minipage}[b]{0.15\linewidth}
{\label{}\includegraphics[width=1\linewidth]{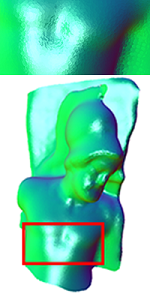}}
\centerline{GLR \cite{zeng20193d}}
\end{minipage}
\begin{minipage}[b]{0.15\linewidth}
{\label{}\includegraphics[width=1\linewidth]{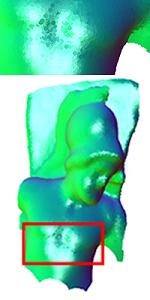}}
\centerline{LPF \cite{digne2017sparse}}
\end{minipage}
\begin{minipage}[b]{0.15\linewidth}
{\label{}\includegraphics[width=1\linewidth]{paper/figures/virtual_scan/soldier/ours.png}}
\centerline{Ours}
\end{minipage} 
\\
\caption{Comparisons on 3 raw scanned models. From top to bottom: David, Relief, and Soldier. }
\label{fig:dne_nonlocal_result}
\end{figure*}

\begin{figure*}[!h]
\centering
\begin{minipage}[b]{0.15\linewidth}
{\label{}\includegraphics[width=1\linewidth]{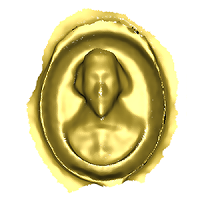}}
\end{minipage}
\begin{minipage}[b]{0.15\linewidth}
{\label{}\includegraphics[width=1\linewidth]{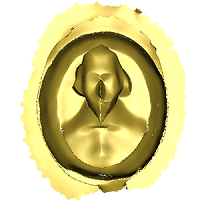}}
\end{minipage}
\begin{minipage}[b]{0.15\linewidth}
{\label{}\includegraphics[width=1\linewidth]{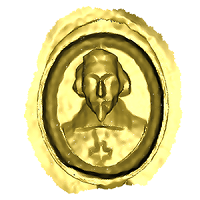}}
\end{minipage}
\begin{minipage}[b]{0.15\linewidth}
{\label{}\includegraphics[width=1\linewidth]{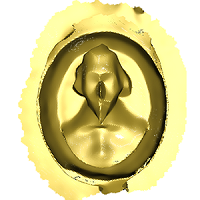}}
\end{minipage}
\begin{minipage}[b]{0.15\linewidth}
{\label{}\includegraphics[width=1\linewidth]{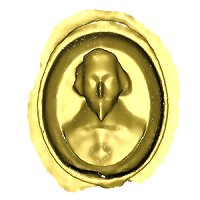}}
\end{minipage}
\begin{minipage}[b]{0.15\linewidth}
{\label{}\includegraphics[width=1\linewidth]{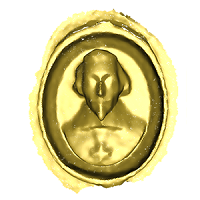}}
\end{minipage} 
\\
\begin{minipage}[b]{0.15\linewidth}
{\label{}\includegraphics[width=1\linewidth]{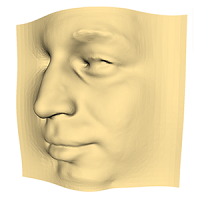}}
\centerline{CLOP \cite{preiner2014continuous}}
\end{minipage}
\begin{minipage}[b]{0.15\linewidth}
{\label{}\includegraphics[width=1\linewidth]{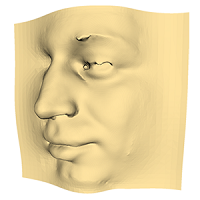}}
\centerline{GPF \cite{lu2017gpf}}
\end{minipage}
\begin{minipage}[b]{0.15\linewidth}
{\label{}\includegraphics[width=1\linewidth]{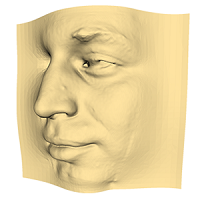}}
\centerline{RIMLS \cite{oztireli2009feature}}
\end{minipage}
\begin{minipage}[b]{0.15\linewidth}
{\label{}\includegraphics[width=1\linewidth]{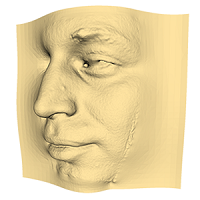}}
\centerline{PointCleanNet \cite{Marie2020PointCleanNet}}
\end{minipage}
\begin{minipage}[b]{0.15\linewidth}
{\label{}\includegraphics[width=1\linewidth]{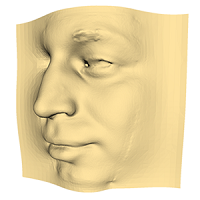}}
\centerline{PointFilter \cite{2020Pointfilter}}
\end{minipage}
\begin{minipage}[b]{0.15\linewidth}
{\label{}\includegraphics[width=1\linewidth]{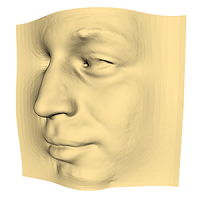}}
\centerline{Ours}
\end{minipage} 
\\
\caption{Surface reconstruction on 2 models. From top to bottom: Richelieu (0.5\% noise), Face (raw noise). }
\label{fig:mesh_reconstruction}
\end{figure*}

\begin{figure*}[!h]
\centering
\begin{minipage}[b]{0.13\linewidth}
{\label{}\includegraphics[width=1\linewidth]{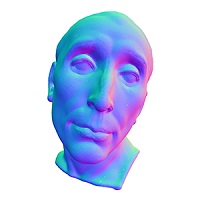}}
\centerline{Nicolo}
\end{minipage}
\begin{minipage}[b]{0.13\linewidth}
{\label{}\includegraphics[width=1\linewidth]{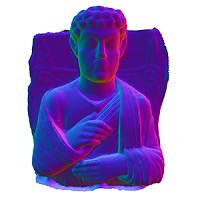}}
\centerline{Funeral man}
\end{minipage}
\begin{minipage}[b]{0.13\linewidth}
{\label{}\includegraphics[width=1\linewidth]{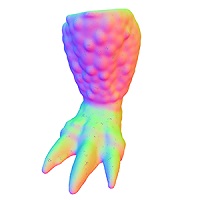}}
\centerline{Leg}
\end{minipage}
\begin{minipage}[b]{0.13\linewidth}
{\label{}\includegraphics[width=1\linewidth]{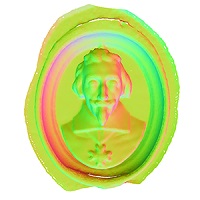}}
\centerline{Richelieu}
\end{minipage}
\begin{minipage}[b]{0.13\linewidth}
{\label{}\includegraphics[width=1\linewidth]{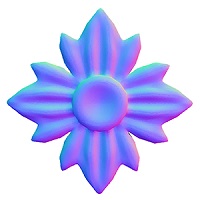}}
\centerline{Flower}
\end{minipage}
\begin{minipage}[b]{0.13\linewidth}
{\label{}\includegraphics[width=1\linewidth]{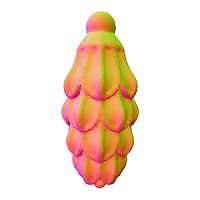}}
\centerline{Fruit}
\end{minipage}
\begin{minipage}[b]{0.13\linewidth}
{\label{}\includegraphics[width=1\linewidth]{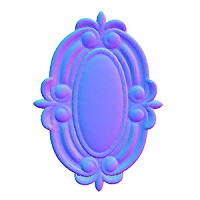}}
\centerline{Mirror}
\end{minipage}
\caption{Ground truths for all synthetic models.}
\label{fig:groundtruth}
\end{figure*}

\textbf{Compared methods.}
We compare our approach with a variety of point cloud filtering techniques which are:
\begin{enumerate}[(1)]
    \item position-based non-learning methods: CLOP \cite{preiner2014continuous}, LPF \cite{digne2017sparse}
    \item normal-based non-learning methods: RIMLS  \cite{oztireli2009feature}, GPF \cite{lu2017gpf}, and GLR \cite{zeng20193d},
    \item position based deep learning methods: PointCleanNet \cite{Marie2020PointCleanNet}, 
    \item normal based deep learning methods: DNP \cite{lu2020deep} and PointFilter \cite{2020Pointfilter},
\end{enumerate}

Notice that PointFilter \cite{2020Pointfilter} utilized ground-truth normal information during the training phase, and the reference phase directly predicts new positions. Thus we classify it to be a normal-based deep learning method.

To make fair comparisons, we employ the rules as follows. (1) We control the local neighborhood of other methods to be in a similar size (if having a local neighborhood parameter) and try our best to tune their involved parameters in obtaining the best visual results. (2) We use EAR \cite{huang2013edge} to upsample all the filtered results by the methods to reach a similar number of points for each model. Regarding surface reconstruction on some models, we use the same parameters for the same model. 

\textbf{Quantitative evaluation metric.} We adopt two quantitative metrics for the evaluation of our method: Chamfer Distance and Mean Square Error.

\begin{enumerate}[(1)]
    \item \textit{Chamfer Distance.}
    The first quantitative evaluation metric we adopt is the Chamfer Distance (CD) which evaluates the distance between two point sets $S_1$ and $S_2$ \cite{2020Pointfilter}. Here, $S_1$ means the ground truth, and $S_2$ indicates the corresponding filtering result. 
    \begin{equation}\label{error_chamferdistance}
        \begin{aligned}
            e_{\mathrm{CD}}\left(S_{1}, S_{2}\right)=\frac{1}{|S_{1}|} \sum_{x \in S_{1}} \min _{y \in S_{2}}\|x-y\|_{2}^{2}+\frac{1}{|S_{2}|} \sum_{y \in S_{2}} \min _{x \in S_{1}}\|y-x\|_{2}^{2}
        \end{aligned}
    \end{equation}
    
    \item \textit{Mean Square Error.}
    We also use Mean Square Error (MSE) as the second quantitative metric \cite{lu2017gpf,2020Pointfilter}.
    \begin{equation}\label{error_mse}
        \begin{aligned}
            e_{\mathrm{MSE}}(S_{1}, S_{2}) = \frac{1}{|S_{1}|} \sum_{x \in S_{1}} \frac{1}{|NN(x)|} \sum_{y\in NN(x)} \|x-y\|_{2}^{2},
        \end{aligned}
    \end{equation}
    where $NN(x)$ denotes the nearest neighbors in $S_2$ of point $x$ in $S_1$. $|NN(x)|=10$, which means we search 10 nearest neighbors for each point $x$ in the ground truth $S_1$.
\end{enumerate}

\subsection{Visual Comparisons}
We conduct experiments on both synthetic point clouds and raw point scans, and the visual results are shown in Figures \ref{fig:synthetic}, \ref{fig:raw}, \ref{fig:dne_nonlocal_result}, and \ref{fig:mesh_reconstruction}. We also provide the parameters used for each model in Table \ref{table:all_parameter}.

\setlength{\tabcolsep}{3pt}
\begin{table}[htb]\tablefont
    \centering
    \caption{The parameters used for each model in Figures \ref{fig:synthetic} and \ref{fig:raw}. }
    \label{table:all_parameter}
    \begin{tabular}{l c c c c c}
        \hline
        \multirow{2}*{Models} & \multirow{2}*{\#.points} & \multicolumn{2}{c}{Parameters} & \multicolumn{2}{c}{Iteration}\\
        & & $K$ & $\theta$ & Scheme & \#. iter.\\
        \hline
        Figure \ref{fig:synthetic} 1st row & 50419 & 100 & 0.05 & 1 & 2 \\
        Figure \ref{fig:synthetic} 2nd row & 67091  & 100 & 0.03 & 1 & 1\\
        Figure \ref{fig:synthetic} 3rd row & 43386 & 100 & 0.2 & 1 & 2\\
        Figure \ref{fig:synthetic} 4th row & 63344  & 120 & 0.05 & 1 & 1\\
        Figure \ref{fig:synthetic} 5th row &  34939 & 80 & 0.1 & 1 & 3\\
        Figure \ref{fig:synthetic} 6th row & 29149  & 80 & 0.07 & 1 & 2\\
        Figure \ref{fig:synthetic} 7th row & 26519  & 100 & 0.05 & 1 & 2\\
        Figure \ref{fig:raw} 1st row & 84398 & 100 & 0.06 & 1 & 3\\
        Figure \ref{fig:raw} 2nd row & 97538 & 120 & 0.05 & 2 & 1\\
        Figure \ref{fig:raw} 3rd row & 54684 & 100 & 0.03 & 1 & 1\\
        Figure \ref{fig:raw} 4th row & 37753 & 80 & 0.05 & 2 & 1\\
        Figure \ref{fig:raw} 5th row & 58197  & 100 & 0.07 & 2 & 1\\
        Figure \ref{fig:raw} 6th row & 35920  & 80 & 0.05 & 1 & 1\\
        \hline
    \end{tabular}
\end{table}

\textbf{Synthetic point clouds.} 
We apply our method to point clouds corrupted with synthetic noise, and compare it with the other five methods. Figure \ref{fig:groundtruth} shows the ground truths of the synthetic models. Figure \ref{fig:synthetic} shows all results on 7 synthetic point clouds contaminated with different levels of noise. 
CLOP \cite{preiner2014continuous} tends to generate overly smoothed results, since it does not consider normal information during denoising. As one of the normal-based methods, GPF \cite{lu2017gpf} often oversharpens some places in the point cloud, which makes the filtering results sometimes unnatural. RIMLS \cite{oztireli2009feature} is a traditional point set projection method which incorporates normal information. Thus it preserves sharp features to some extent (but sometimes overshapening certain area). These methods are all non-deep-learning methods. Now let us take a look at the results by two deep learning methods. PointCleanNet \cite{Marie2020PointCleanNet} does not consider normal information, but it can retain details to certain extent. This is mainly because it utilizes a constrained $L_2$ loss function. Also, it is sometimes difficult for this method to completely remove noise, resulting in the retaining of certain noise in the filtering results. As for our method, it is a completely position-based method. In other words, it does not take account of normal information. From Figure \ref{fig:synthetic}, we can see that it outperforms the position-based methods (i.e. CLOP), and is comparable to or even better than the normal-based methods (i.e. GPF, RIMLS, PointCleanNet, PointFilter). In specific, our method is better than GPF in all point cloud models. In the first row (i.e. first model), our method produces better results than all other methods, in terms of preserving sharp features and smoothing point set surface. In the second and third rows (i.e. the second and third models), the results by our method are comparable to those by PointFilter. In the fourth row, our method generates a better result than PointFilter, and it is comparable to PointCleanNet. In the fifth, sixth, and seventh rows, the proposed method and PointFilter generate very similar filtering results. While PointFilter utilizes normal information for training, our method purely depends on positional information (i.e. point positions).

\begin{figure*}[htbp]
\centering
\begin{minipage}[b]{0.9\linewidth}
\begin{minipage}[b]{0.16\linewidth}
{\label{}\includegraphics[width=1\linewidth]{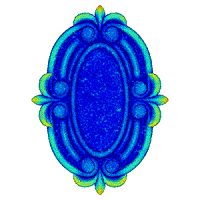}}
\end{minipage}
\begin{minipage}[b]{0.16\linewidth}
{\label{}\includegraphics[width=1\linewidth]{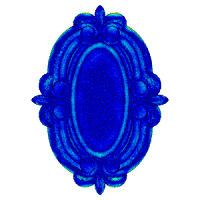}}
\end{minipage}
\begin{minipage}[b]{0.16\linewidth}
{\label{}\includegraphics[width=1\linewidth]{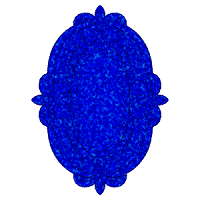}}
\end{minipage}
\begin{minipage}[b]{0.16\linewidth}
{\label{}\includegraphics[width=1\linewidth]{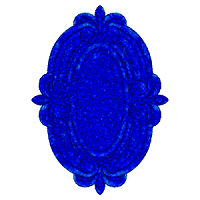}}
\end{minipage}
\begin{minipage}[b]{0.16\linewidth}
{\label{}\includegraphics[width=1\linewidth]{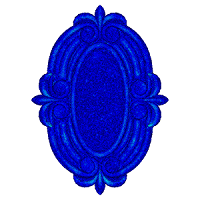}}
\end{minipage} 
\begin{minipage}[b]{0.16\linewidth}
{\label{}\includegraphics[width=1\linewidth]{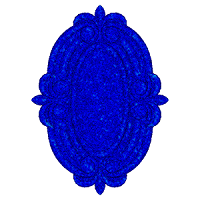}}
\end{minipage} 
\\
\begin{minipage}[b]{0.16\linewidth}
{\label{}\includegraphics[width=1\linewidth]{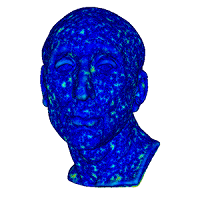}}
\centerline{CLOP \cite{preiner2014continuous} }
\end{minipage}
\begin{minipage}[b]{0.16\linewidth}
{\label{}\includegraphics[width=1\linewidth]{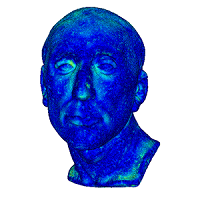}}
\centerline{GPF \cite{lu2017gpf}}
\end{minipage}
\begin{minipage}[b]{0.16\linewidth}
{\label{}\includegraphics[width=1\linewidth]{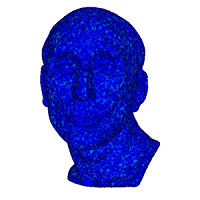}}
\centerline{RIMLS \cite{oztireli2009feature}}
\end{minipage}
\begin{minipage}[b]{0.16\linewidth}
{\label{}\includegraphics[width=1\linewidth]{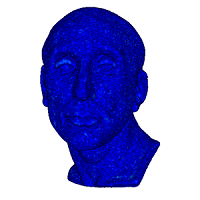}}
\centerline{PointCleanNet \cite{Marie2020PointCleanNet}}
\end{minipage}
\begin{minipage}[b]{0.16\linewidth}
{\label{}\includegraphics[width=1\linewidth]{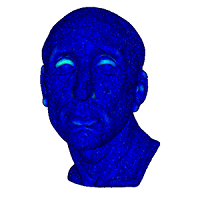}}
\centerline{PointFilter \cite{2020Pointfilter}}
\end{minipage}
\begin{minipage}[b]{0.16\linewidth}
{\label{}\includegraphics[width=1\linewidth]{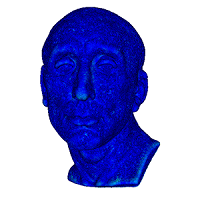}}
\centerline{Ours}
\end{minipage} 
\end{minipage}
\begin{minipage}[b]{0.08\linewidth}
{\label{}\includegraphics[width=1\linewidth]{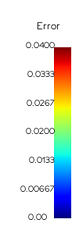}}
\end{minipage} 
\caption{ Visualization in Mean Square Error (MSE). The overall errors ($\times10^{-3}$) for different methods over two models are shown in the figure.}
\label{fig:visual_mse}
\end{figure*}

\begin{table*}[htb]
    \centering
    \caption{ Quantitative evaluation results of the compared methods and our method on the synthetic point clouds in Figure \ref{fig:synthetic}. Note that * represents deep learning methods. Chamfer Distance ($\times10^{-5}$) are used here. The three best methods for each model are highlighted. }
    \label{table:errors_evaluation_chamfer_syn}
    \begin{tabular}{l c c c c c c c c c c}
        \hline
        Methods & \tabincell{c}{Figure \ref{fig:synthetic}\\1st row} & \tabincell{c}{Figure \ref{fig:synthetic}\\2nd row} & \tabincell{c}{Figure \ref{fig:synthetic}\\3rd row} & \tabincell{c}{Figure \ref{fig:synthetic}\\4th row} & \tabincell{c}{Figure \ref{fig:synthetic}\\5th row} & \tabincell{c}{Figure \ref{fig:synthetic}\\6th row} & \tabincell{c}{Figure \ref{fig:synthetic}\\7th row} & Avg.\\ 
        \hline
        CLOP \cite{preiner2014continuous}            & 6.08          & 6.02           & \textbf{5.07} & 9.42          & 15.51          & 9.16                 & 17.01 & 9.75\\
        GPF \cite{lu2017gpf}                         & 12.49         & 26.57          & 23.83         & 29.50         & 12.79          & 6.47                 & 14.77 & 18.06\\
        RIMLS \cite{oztireli2009feature}             & 10.85         & 2.48           & 14.83         & \textbf{2.96} & 10.38          & \textbf{2.90}        & 7.24 & 7.38\\
        PointCleanNet* \cite{Marie2020PointCleanNet} & \textbf{2.87} & \textbf{2.05}  & 5.62          & \textbf{2.03} & \textbf{3.62}  & \textbf{3.69}        & \textbf{5.35} & \textbf{3.60}\\
        PointFilter* \cite{2020Pointfilter}          & \textbf{2.38} & \textbf{1.66}  & \textbf{3.57} & \textbf{1.89} & \textbf{3.22}  & \textbf{4.46}        & \textbf{4.21} & \textbf{3.06}\\
        \hline
        Ours                                         & \textbf{3.28} & \textbf{2.33}  & \textbf{4.93} & 6.89          & \textbf{4.73}  & 4.50                 & \textbf{4.20} & \textbf{4.41}\\
        \hline
    \end{tabular}
\end{table*}

\begin{table*}[!h]
    \centering
    \caption{ Quantitative evaluation results of the compared methods and our method on the synthetic point clouds in Figure \ref{fig:synthetic}. Note that * represents deep learning methods. Mean Square Error ($\times10^{-3}$) are used here. The three best methods for each model are highlighted. }
    \label{table:errors_evaluation_meansqure_syn}
    \begin{tabular}{l c c c c c c c c c c}
        \hline
        Methods & \tabincell{c}{Figure \ref{fig:synthetic}\\1st row} & \tabincell{c}{Figure \ref{fig:synthetic}\\2nd row} & \tabincell{c}{Figure \ref{fig:synthetic}\\3rd row} & \tabincell{c}{Figure \ref{fig:synthetic}\\4th row} & \tabincell{c}{Figure \ref{fig:synthetic}\\5th row} & \tabincell{c}{Figure \ref{fig:synthetic}\\6th row} & \tabincell{c}{Figure \ref{fig:synthetic}\\7th row} & Avg.\\  
        \hline
        CLOP \cite{preiner2014continuous}            & 7.50          & 6.36           & \textbf{8.89} & 7.54          & 9.52          & 8.86            & 8.62 & 8.18\\
        GPF \cite{lu2017gpf}                         & 9.11          & 10.53          & 12.80         & 11.51         & 9.77          & 8.74            & 9.35 & 10.26\\
        RIMLS \cite{oztireli2009feature}             & 9.94          & 5.83           & 12.25         & \textbf{6.35} & 10.09         & \textbf{8.13}   & 8.80 & 8.77\\
        PointCleanNet* \cite{Marie2020PointCleanNet} & \textbf{6.89} & \textbf{5.59}  & 9.52          & \textbf{5.89} & \textbf{7.28} & \textbf{8.18}   & \textbf{7.54} & \textbf{7.27}\\
        PointFilter* \cite{2020Pointfilter}          & \textbf{6.21} & \textbf{5.25}  & \textbf{8.82} & \textbf{5.68} & \textbf{6.97} & \textbf{8.15}   & \textbf{6.92} & \textbf{6.86}\\
        \hline
        Ours                                         & \textbf{6.66}  & \textbf{5.53}  & \textbf{9.20} & 7.43          & \textbf{7.61} & 8.38            & \textbf{7.28} & \textbf{7.44}\\
        \hline
    \end{tabular}
\end{table*}

\begin{table}[htb]
    \centering
    \caption{ Quantitative evaluation results of the compared methods and our method on some scanned point clouds (with ground truth) in Figure \ref{fig:raw}. Note that * represents deep learning methods.  Chamfer Distance ($\times10^{-3}$) are used here. The three best methods for each model are highlighted. }
    \label{table:errors_evaluation_chamfer_vir}
    \begin{tabular}{l c c c}
        \hline
        Methods & \tabincell{c}{Figure \ref{fig:raw}\\4th row} &  \tabincell{c}{Figure \ref{fig:raw}\\5th row} & Avg.\\
        \hline
        CLOP \cite{preiner2014continuous}              & 1.13          & \textbf{1.42} & 1.28 \\
        GPF \cite{lu2017gpf}                           & \textbf{1.06} & 1.63          & 1.35 \\
        RIMLS \cite{oztireli2009feature}               & \textbf{0.97} & 1.45          & \textbf{1.21} \\
        PointCleanNet* \cite{Marie2020PointCleanNet}   & \textbf{0.98} & \textbf{1.42} & \textbf{1.20} \\
        PointFilter* \cite{2020Pointfilter}            & 1.07          & 1.47          & 1.27 \\
        \hline
        Ours                                           & \textbf{1.06} & \textbf{1.32} & \textbf{1.19} \\
        \hline
    \end{tabular}
\end{table}

\begin{table}[htb]
    \centering
    \caption{ Quantitative evaluation results of the compared methods and our method on some scanned point clouds (with ground truth) in Figure \ref{fig:raw}. Note that * represents deep learning methods.  Mean Square Error ($\times10^{-2}$) are used here. The three best methods for each model are highlighted. }
    \label{table:errors_evaluation_mse_vir}
    \begin{tabular}{l c c c}
        \hline
        Methods & \tabincell{c}{Figure \ref{fig:raw}\\4th row} &  \tabincell{c}{Figure \ref{fig:raw}\\5th row} & Avg.\\
        \hline
        CLOP \cite{preiner2014continuous}            & \textbf{1.99} & \textbf{1.50} & \textbf{1.75} \\
        GPF \cite{lu2017gpf}                         & 2.05          & 1.62          & 1.84 \\
        RIMLS \cite{oztireli2009feature}             & 2.00          & \textbf{1.52} & \textbf{1.76} \\
        PointCleanNet* \cite{Marie2020PointCleanNet} & \textbf{1.99} & \textbf{1.52} & \textbf{1.76} \\
        PointFilter* \cite{2020Pointfilter}          & 2.00          & 1.53          & 1.77 \\
        \hline
        Ours                                         & \textbf{1.90} & \textbf{1.48} & \textbf{1.69} \\
        \hline
    \end{tabular}
\end{table}

\begin{figure*}[htb]
\centering
\begin{minipage}[b]{0.16\linewidth}
{\label{}\includegraphics[width=1\linewidth]{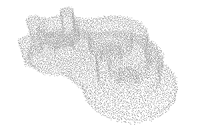}}
\centerline{Noisy input (points: 20k)}
\end{minipage}
\begin{minipage}[b]{0.16\linewidth}
{\label{}\includegraphics[width=1\linewidth]{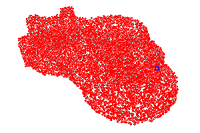}}
\centerline{$K$ = 10}
\end{minipage}
\begin{minipage}[b]{0.16\linewidth}
{\label{}\includegraphics[width=1\linewidth]{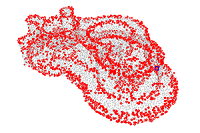}}
\centerline{$K$ = 30}
\end{minipage}
\begin{minipage}[b]{0.16\linewidth}
{\label{}\includegraphics[width=1\linewidth]{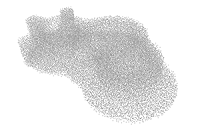}}
\centerline{Noisy input (points: 50k)}
\end{minipage}
\begin{minipage}[b]{0.16\linewidth}
{\label{}\includegraphics[width=1\linewidth]{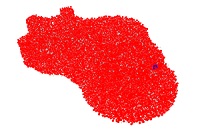}}
\centerline{$K$ = 30}
\end{minipage}
\begin{minipage}[b]{0.16\linewidth}
{\label{}\includegraphics[width=1\linewidth]{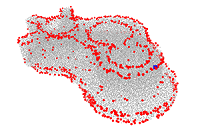}}
\centerline{$K$ = 50}
\end{minipage}
\\
\vspace{0.5cm}
\begin{minipage}[b]{0.16\linewidth}
{\label{}\includegraphics[width=1\linewidth]{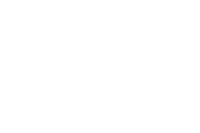}}
\end{minipage}
\begin{minipage}[b]{0.16\linewidth}
{\label{}\includegraphics[width=1\linewidth]{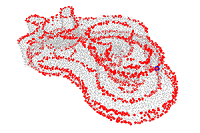}}
\centerline{$K$ = \textbf{50}}
\end{minipage}
\begin{minipage}[b]{0.16\linewidth}
{\label{}\includegraphics[width=1\linewidth]{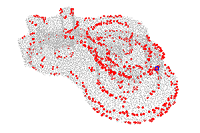}}
\centerline{$K$ = 100}
\end{minipage}
\begin{minipage}[b]{0.16\linewidth}
{\label{}\includegraphics[width=1\linewidth]{paper/figures/ablation/parameter/blank.png}}
\end{minipage}
\begin{minipage}[b]{0.16\linewidth}
{\label{}\includegraphics[width=1\linewidth]{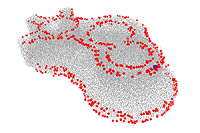}}
\centerline{$K$ = \textbf{100}}
\end{minipage}
\begin{minipage}[b]{0.16\linewidth}
{\label{}\includegraphics[width=1\linewidth]{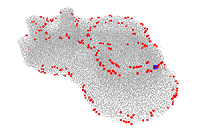}}
\centerline{$K$ = 200}
\end{minipage}
\caption{Ablation study on $K$ ($\theta = 0.1$). Blue point is the centred point of the given patch. Red points are the central points of the similar patches according to the given patch. }
\label{fig:parameter_k}
\end{figure*}

\textbf{Raw point scans.}
In addition to the above experiments on synthetic point clouds, we also compare our method with those methods on raw point scans. As shown in Figure \ref{fig:raw}, there are 6 different raw point scans. From the figure, we can easily observe that the proposed method produces very good results. In general, it generates the best filtering outcomes among all methods, in terms of preserving sharp features as well as smoothing out the noise. To be specific, in the first model, our method better preserves the eyelid area while eliminating noise. In the second model, our method enables a great smoothing result while preserving the decoration on the hat. In the third model, our method preserves the curved lines of the relief while smoothing the point set surface. In the fourth model, our method preserves the nose better and naturally. In the fifth model, our method is better at removing noise while preserving sharp features. In the sixth model, our method better preserves the chest muscle area than other methods. By contrast, other methods usually produce either over smoothed filtering results, oversharpened filtering results, or noise-retaining outcomes. Notice that the third, fifth, and sixth models are relief-like models which often share many similar patterns in the point cloud. Figure \ref{fig:mesh_reconstruction} shows the surface reconstruction of different methods on two models. We can see from the figure that our method preserves features better than most compared methods in the first row. As for the second row, our method has better smoothing effect than other methods on cheek area and preserves features well in nose and eyelid. RIMLS often retains noise to some extent. In addition, we also compare our method with DNP \cite{lu2020deep}, GLR \cite{zeng20193d}, and LPF \cite{digne2017sparse}. As shown in Figure \ref{fig:dne_nonlocal_result}, our approach still achieves very competitive filtering results, without utilizing normal  information. In comparison, the results of DNP \cite{lu2020deep} and GLR \cite{zeng20193d} certainly benefit from their used normal information. The good performance of our method is mainly due to the proper design of the similarity metric and the full utilization of non-local similar information.

\subsection{Quantitative Comparisons}
We also compare the filtering results of all methods quantitatively. As mentioned above, we use two metrics for quantitative evaluation, which are Chamfer Distance and Mean Square Error. Tables \ref{table:errors_evaluation_chamfer_syn} and \ref{table:errors_evaluation_meansqure_syn} show the quantitative comparisons for the methods on the synthetic point clouds. In general, PointFilter \cite{2020Pointfilter} induces the smallest Chamfer Distance and Mean Square Error. It is not surprising to see this as PointFilter is a deep learning method that relies on a large amount of data for training. As a non-deep-learning method that only utilizes positional information, it is interesting to observe that our method generates close errors to PointFilter, and often ranks the second or the third place. We also visualize the MSE errors on two models in Figure \ref{fig:visual_mse}.

Tables \ref{table:errors_evaluation_chamfer_vir} and \ref{table:errors_evaluation_mse_vir} show the error comparisons on the scanned data with ground truth. As for the Chamfer Distance, our method ranks the third and the best, respectively. With regard to the Mean Square Error, our method achieves the  best on all two raw point scans. This also reflects the effectiveness and superiority of the proposed method.

\subsection{Ablation Studies}

\textbf{Parameters.}
We test different $K$ on two point clouds ($\sim$20k and $\sim$50k). The searched similar patches are shown in Figure \ref{fig:parameter_k}. If $K$ is set to a very small number, the local neighborhood is very small, which makes it hard to contain sufficiently distinct information. As a result, many similar patches can be found. When $K$ increases, the number of similar patches reduces. If $K$ is too large, it will take much more time to find similar patches. We typically set a larger $K$ for a point cloud with more points, which will allow a local patch to include sufficiently distinct information for effective similar patches finding.

\begin{figure}[htb]
\centering
\begin{minipage}[b]{0.32\linewidth}
{\label{}\includegraphics[width=1\linewidth]{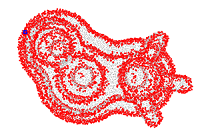}}
\centerline{$\theta$ = 0.2}
\end{minipage}
\begin{minipage}[b]{0.32\linewidth}
{\label{}\includegraphics[width=1\linewidth]{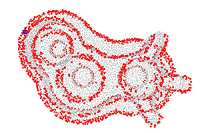}}
\centerline{$\theta$ = \textbf{0.1}}
\end{minipage}
\begin{minipage}[b]{0.32\linewidth}
{\label{}\includegraphics[width=1\linewidth]{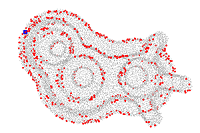}}
\centerline{$\theta$ = 0.05}
\end{minipage} 
\caption{ Ablation study on $\theta$. This model has $\sim$20k points, and we set $K = 50$. Bluepoint is the central point of the given patch. Red points are the central points of similar patches according to the given patch. }
\label{fig:parameter_theta}
\end{figure}

As for the threshold $\theta$, we test different values and show the searched similar patches in Figure \ref{fig:parameter_theta}. We can see that a larger $\theta$ loses the constraint and allows more patches to be included as similar patches. In comparison, a smaller $\theta$ tightens up the constraint, which induces a smaller number of similar patches.
We also experimented with parameters for some different models, and Table \ref{table:similar_numbers} shows the number of similar patches we found with different parameters. 

\setlength{\tabcolsep}{3pt}
\begin{table}[!h]\tablefont
    \centering
    \caption{The number of similar patches found using different parameters. The parameters for each model we used are highlighted. }
    \label{table:similar_numbers}
    \begin{tabular}{l c c c c}
        \hline
        \multirow{2}*{Models} & \multirow{2}*{\#. points} & \#. similar & \multicolumn{2}{c}{Parameters}\\
                              &                           & patches     & $K$ & $\theta$\\
        \hline
        \multirow{6}*{\tabincell{l}{Figure \ref{fig:parameter_k} ($\sim$20k points)\\ and Figure \ref{fig:parameter_theta}}} & \multirow{6}*{19398} & 13225  & 10 & 0.1\\
        & & 3531  & 30 & 0.1\\
        & & \textbf{2257}  & \textbf{50} & \textbf{0.1}\\
        & & 971  & 100 & 0.1\\
        & & 6241  & 50 & 0.2\\
        & & 685  & 50 & 0.05\\
        \hline
        \multirow{4}*{Figure \ref{fig:parameter_k} ($\sim$50k points)} & \multirow{4}*{51745}   & 41126  & 30 & 0.1\\
        & & 1249  & 50 & 0.1\\
        & & \textbf{726}  & \textbf{100} & \textbf{0.1}\\
        & & 319  & 200 & 0.1\\
        \hline
        \multirow{5}*{Figure \ref{fig:synthetic} 1st row} & \multirow{5}*{50419}  & 8027  & 50 & 0.05\\
        & & \textbf{2036}  & \textbf{100} & \textbf{0.05}\\
        & & 178  & 150 & 0.05\\
        & & 6079  & 100 & 0.1\\
        & & 383  & 100 & 0.02\\
        \hline
        \multirow{5}*{Figure \ref{fig:synthetic} 6th row} & \multirow{5}*{29149}  & 7798  & 50 & 0.07\\
        & & \textbf{3940}  & \textbf{80} & \textbf{0.07}\\
        & & 2015  & 120 & 0.07\\
        & & 7275  & 80 & 0.1\\
        & & 1388  & 80 & 0.04\\
        \hline
        \multirow{5}*{Figure \ref{fig:raw} 6th row} & \multirow{5}*{35920}  & 3903 & 50 & 0.05\\
        & & \textbf{1709}  & \textbf{80} & \textbf{0.05}\\
        & & 1248  & 120 & 0.05\\
        & & 5189 & 80 & 0.1\\
        & & 235  & 80 & 0.02\\
        \hline
    \end{tabular}
\end{table}

\begin{figure}[!h]
\centering
\begin{minipage}[b]{0.32\linewidth}
\subfigure[1 iteration]{\label{fig:iter_num_a}\includegraphics[width=1\linewidth]{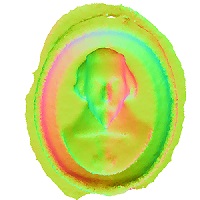}}
\end{minipage}
\begin{minipage}[b]{0.32\linewidth}
\subfigure[2 iterations]{\label{fig:iter_num_b}\includegraphics[width=1\linewidth]{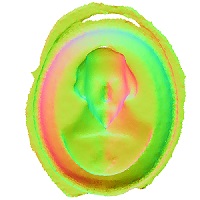}}
\end{minipage}
\begin{minipage}[b]{0.32\linewidth}
\subfigure[3 iterations]{\label{fig:iter_num_c}\includegraphics[width=1\linewidth]{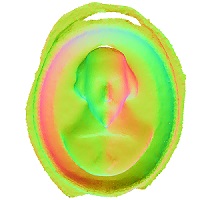}}
\end{minipage} 
\\
\begin{minipage}[b]{0.32\linewidth}
\subfigure[1 iteration]{\label{fig:iter_num_d}\includegraphics[width=1\linewidth]{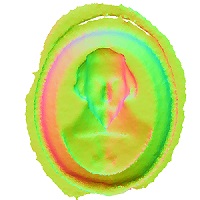}}
\end{minipage}
\begin{minipage}[b]{0.32\linewidth}
\subfigure[2 iterations]{\label{fig:iter_num_e}\includegraphics[width=1\linewidth]{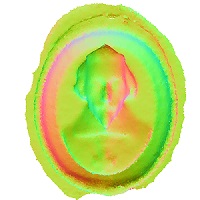}}
\end{minipage}
\begin{minipage}[b]{0.32\linewidth}
\subfigure[3 iterations]{\label{fig:iter_num_f}\includegraphics[width=1\linewidth]{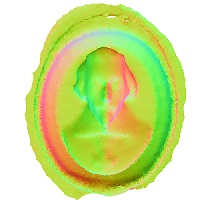}}
\end{minipage} 
\caption{ Results of different iterations using the two iteration schemes. (a)-(c) Results of one, two, and three iterations using the first iteration scheme. (d)-(f) Results of one, two, and three iterations using the second iteration scheme. }
\label{fig:iter_num}
\end{figure}

With respect to the number of iterations, it is intuitive that more iterations will enable smoother filtering outcomes. We show an example in Figure \ref{fig:iter_num}, and the errors are shown in Table \ref{table:errors_evaluation_iter}.

\begin{table}[!h]
    \centering
    \caption{Quantitative evaluation results of the different iterations using the two iteration schemes in Figure \ref{fig:iter_num}. Chamfer Distance ($\times10^{-5}$) and Mean Square Error ($\times10^{-3}$) are used here.}
    \label{table:errors_evaluation_iter}
    \begin{tabular}{l c c}
        \hline
        Models & Chamfer Distance &  Mean Square Error \\ 
        \hline
        Figure \ref{fig:iter_num_a} & 6.889 & 7.431\\
        Figure \ref{fig:iter_num_b} & 16.294 & 9.158\\
        Figure \ref{fig:iter_num_c} & 22.681 & 10.056\\
        Figure \ref{fig:iter_num_d} & 6.889 & 7.431\\
        Figure \ref{fig:iter_num_e} & 6.896 & 7.426\\
        Figure \ref{fig:iter_num_f} & 6.899 & 7.423\\
        \hline
    \end{tabular}
\end{table}

\begin{figure}[htp]
\centering
\begin{minipage}[b]{0.32\linewidth}
{\label{}\includegraphics[width=1\linewidth]{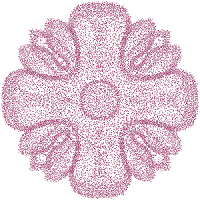}}
\centerline{Noisy input}
\end{minipage}
\begin{minipage}[b]{0.32\linewidth}
{\label{}\includegraphics[width=1\linewidth]{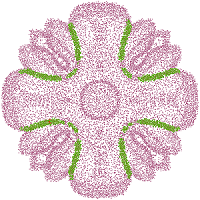}}
\centerline{Non-local method}
\end{minipage}
\begin{minipage}[b]{0.32\linewidth}
{\label{}\includegraphics[width=1\linewidth]{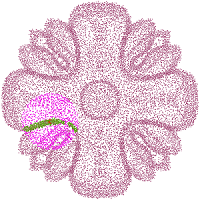}}
\centerline{Local method}
\end{minipage} 
\caption{Non-local method versus local method. The points with bright pink show the local neighborhood range of the local method. }
\label{fig:nonlocal_local}
\end{figure}

\textbf{Non-local versus local.}
We also compare the choices of using the non-local method and the local method to find similar patches. For a local patch $M$, in the non-local method, we compare patch $M$ with patches in the whole point cloud to determine whether they are similar or not. In contrast, in the local method we define a larger local neighborhood based on the size of patch $M$ (e.g. three times the local patch size), and only calculate the similarity for the patches in this range to find the similar patches of patch $M$. The similar patches finding results of our non-local method and the local method are illustrated in Figure \ref{fig:nonlocal_local}. 

We observed that the local scheme can only search very limited similar patches, while our non-local method can search similar patches in the whole point cloud and can provide sufficient information in point cloud filtering.

\begin{figure}[!h]
\centering
\begin{minipage}[b]{0.32\linewidth}
{\label{}\includegraphics[width=1\linewidth]{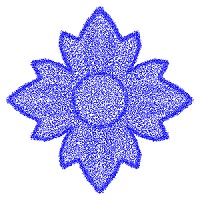}}
\centerline{Noisy input}
\end{minipage}
\begin{minipage}[b]{0.32\linewidth}
{\label{}\includegraphics[width=1\linewidth]{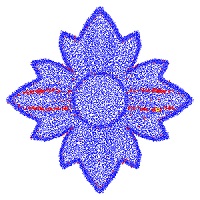}}
\centerline{Covariance matrix}
\end{minipage}
\begin{minipage}[b]{0.32\linewidth}
{\label{}\includegraphics[width=1\linewidth]{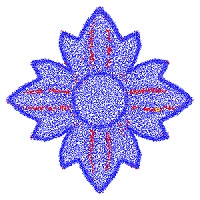}}
\centerline{Ours}
\end{minipage} 
\caption{Using covariance matrix $D$ for searching similar patches. Red points are the central points of the found similar patches. }
\label{fig:cov_rpca}
\end{figure}

\begin{figure}[!h]
\centering
\begin{minipage}[c]{0.1\linewidth}
\begin{center}
Noisy input
\end{center}
\end{minipage}
\begin{minipage}[c]{0.28\linewidth}
{\label{}\includegraphics[width=1\linewidth]{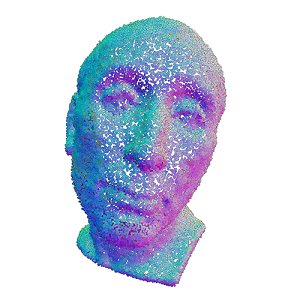}}
\end{minipage}
\begin{minipage}[c]{0.28\linewidth}
{\label{}\includegraphics[width=1\linewidth]{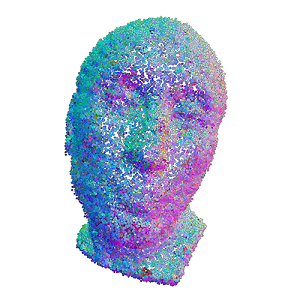}}
\end{minipage}
\begin{minipage}[c]{0.28\linewidth}
{\label{}\includegraphics[width=1\linewidth]{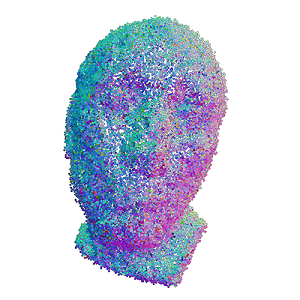}}
\end{minipage} 
\\
\begin{minipage}[c]{0.1\linewidth}
\begin{center}
CLOP \cite{preiner2014continuous}
\end{center}
\end{minipage}
\begin{minipage}[c]{0.28\linewidth}
{\label{}\includegraphics[width=1\linewidth]{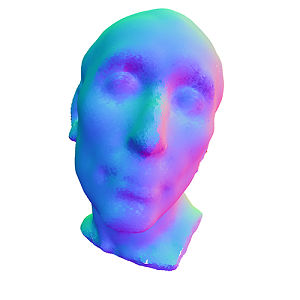}}
\end{minipage}
\begin{minipage}[c]{0.28\linewidth}
{\label{}\includegraphics[width=1\linewidth]{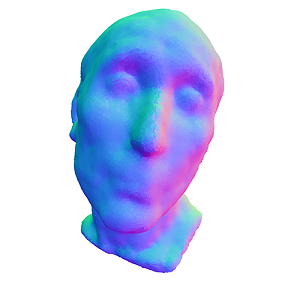}}
\end{minipage}
\begin{minipage}[c]{0.28\linewidth}
{\label{}\includegraphics[width=1\linewidth]{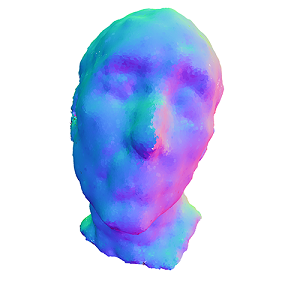}}
\end{minipage} 
\\
\begin{minipage}[c]{0.1\linewidth}
\begin{center}
\footnotesize{PointCl- eanNet} \normalsize{\cite{Marie2020PointCleanNet}}
\end{center}
\end{minipage}
\begin{minipage}[c]{0.28\linewidth}
{\label{}\includegraphics[width=1\linewidth]{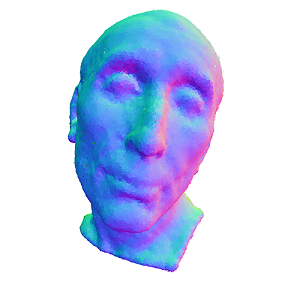}}
\end{minipage}
\begin{minipage}[c]{0.28\linewidth}
{\label{}\includegraphics[width=1\linewidth]{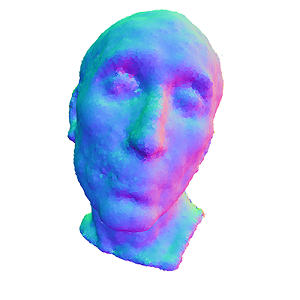}}
\end{minipage}
\begin{minipage}[c]{0.28\linewidth}
{\label{}\includegraphics[width=1\linewidth]{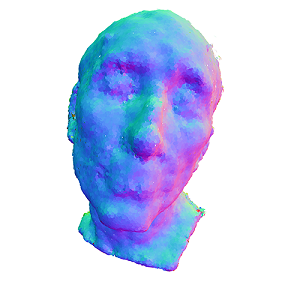}}
\end{minipage} 
\\
\begin{minipage}[c]{0.1\linewidth}
\begin{center}
Ours
\end{center}
\end{minipage}
\begin{minipage}[c]{0.28\linewidth}
{\label{}\includegraphics[width=1\linewidth]{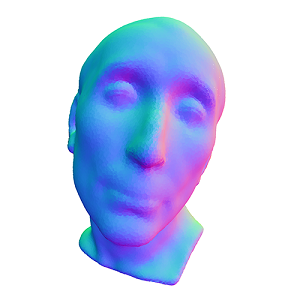}}
\end{minipage}
\begin{minipage}[c]{0.28\linewidth}
{\label{}\includegraphics[width=1\linewidth]{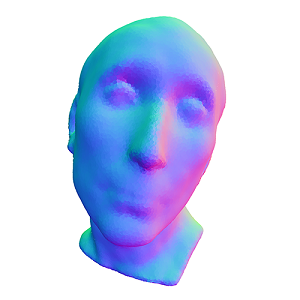}}
\end{minipage}
\begin{minipage}[c]{0.28\linewidth}
{\label{}\includegraphics[width=1\linewidth]{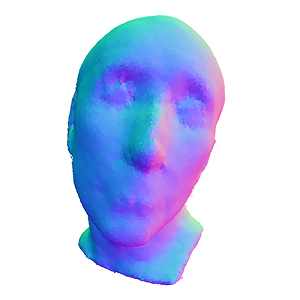}}
\end{minipage} 
\\
\begin{minipage}[c]{0.1\linewidth}
\begin{center}
\end{center}
\end{minipage}
\begin{minipage}[c]{0.28\linewidth}
\centerline{0.5\% noise}
\end{minipage}
\begin{minipage}[c]{0.28\linewidth}
\centerline{1.0\% noise}
\end{minipage}
\begin{minipage}[c]{0.28\linewidth}
\centerline{1.5\% noise}
\end{minipage} 
\caption{Filtering results of models with different levels of noise. }
\label{fig:noise_level}
\end{figure}

\textbf{Covariance matrix for searching similar patches.} An alternative way to search similar patches of a given patch is using the covariance matrix $D$ defined in Section \ref{sec::Position_Update}. In other words, the eigenvalues of $D$ form a vector, and the similarity can be calculated as the distance of two such vectors of two patches. As shown in Figure \ref{fig:cov_rpca}, this strategy produces poor results as it misses many similar patches. In contrast, our method is more robust in finding similar patches in the presence of noise.

\textbf{Noise level.} 
In addition to the above study, we also test different levels of noise with other methods to show the robustness of our method. 
Figure \ref{fig:noise_level} shows the visual comparison and Table \ref{table:errors_evaluation_noise} indicates the errors of them. From the above figure and table, our method can still get good denoising results while PointCleanNet and CLOP do not remove noise well when the noise becomes larger. Since our approach is position-based, it is capable of handling large noise. However, features are becoming hard to retain with increasing levels of noise. This is mainly because larger noise destroys feature information more severely (deviating largely from the underlying surfaces), and the found similar patches provide less useful information relevant to features. In fact, nearly all existing methods can hardly handle this. Thanks to the non-local and positional properties, our method can still filter out the noise and smooth the point set surface effectively. 

\begin{table}[!h]
\vspace{0.2cm}
    \centering
    \caption{Quantitative evaluation results of models with different levels of noise in Figure \ref{fig:noise_level}.  Chamfer Distance ($\times10^{-5}$) and Mean Square Error ($\times10^{-3}$) are used here. The best method for each model is highlighted.}
    \label{table:errors_evaluation_noise}
    \begin{tabular}{l c c c c c}
        \hline
        Methods & Metrics & \tabincell{c}{0.5\% \\noise} &  \tabincell{c}{1.0\% \\noise} &  \tabincell{c}{1.5\% \\noise} & Avg.\\
        \hline
        CLOP \cite{preiner2014continuous}           & \multirow{3}*{CD}   &  5.56 & 6.08 & 8.35 & 6.66\\
        PointCleanNet \cite{Marie2020PointCleanNet} & &  \textbf{2.10} & \textbf{2.87} & 14.23 & 6.40\\ 
        Ours                                        & & 2.65 & 3.28 & \textbf{6.43} & \textbf{4.12}\\
        \hline
        CLOP \cite{preiner2014continuous}           & \multirow{3}*{MSE} & 7.36 & 7.50 & 8.05 & 7.64\\
        PointCleanNet \cite{Marie2020PointCleanNet} & & 6.43 & 6.89 & 10.36 & 7.89\\ 
        Ours                                        & & \textbf{6.39} & \textbf{6.66} & \textbf{7.57} & \textbf{6.84}\\
        \hline
    \end{tabular}
\end{table}

\begin{figure}[htb]
\centering
\begin{minipage}[b]{0.32\linewidth}
{\label{}\includegraphics[width=1\linewidth]{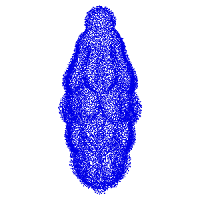}}
\centerline{40\% points}
\end{minipage}
\begin{minipage}[b]{0.32\linewidth}
{\label{}\includegraphics[width=1\linewidth]{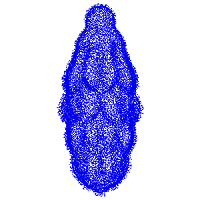}}
\centerline{60\% points}
\end{minipage}
\begin{minipage}[b]{0.32\linewidth}
{\label{}\includegraphics[width=1\linewidth]{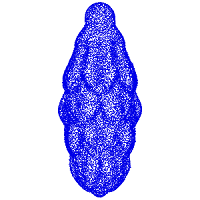}}
\centerline{100\% points}
\end{minipage} 
\caption{Filtering results on different sampling levels.}
\label{fig:selected_denoise}
\end{figure}

\textbf{Sampling level.
We conduct experiments at different sampling levels. It is clear from Figure \ref{fig:selected_denoise} that the best results were obtained by choosing 100\% of the points, although it takes the longest time (Table \ref{table:runtime_pointslevel}). In essence, sampling is a tradeoff strategy between speed and denoising quality.}

\setlength{\tabcolsep}{3pt}
\begin{table}[htb]\tablefont
    \centering
    \caption{Runtime performance (in seconds) for different sampling levels.}
    \label{table:runtime_pointslevel}
    \begin{tabular}{l c c c}
        \hline
        Models & $40\%$ points & $60\%$ points & $100\%$ points\\
        \hline
        Ours (Scheme 1) & 283 & 423 & 793 \\
        Ours (Scheme 2) & 260 & 467 & 840 \\
        \hline
    \end{tabular}
\end{table}

\begin{figure}[htb]
\centering
\begin{minipage}[b]{0.3\linewidth}
{\label{}\includegraphics[width=1\linewidth]{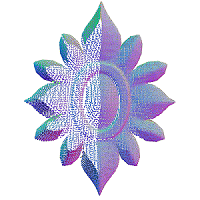}}
\centerline{Ground truth}
\end{minipage}
\begin{minipage}[b]{0.3\linewidth}
{\label{}\includegraphics[width=1\linewidth]{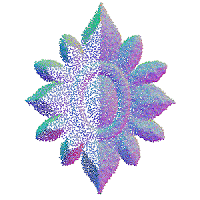}}
\centerline{Noisy input}
\end{minipage}
\\
\begin{minipage}[b]{0.3\linewidth}
{\label{}\includegraphics[width=1\linewidth]{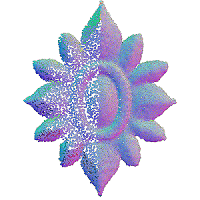}}
\centerline{PointCleanNet \cite{Marie2020PointCleanNet}}
\end{minipage}
\begin{minipage}[b]{0.3\linewidth}
{\label{}\includegraphics[width=1\linewidth]{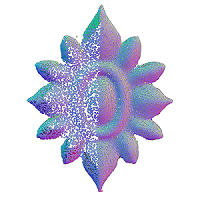}}
\centerline{PointFilter \cite{2020Pointfilter}}
\end{minipage}
\begin{minipage}[b]{0.3\linewidth}
{\label{}\includegraphics[width=1\linewidth]{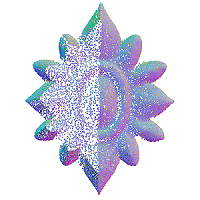}}
\centerline{Ours}
\end{minipage} 
\caption{Filtering result of PointCleanNet \cite{Marie2020PointCleanNet}, PointFilter \cite{2020Pointfilter}  and our method over the Decoration point cloud with irregular sampling. }
\label{fig:uneven}
\end{figure}

\textbf{Irregular sampling.} 
We conduct experiments on a model with irregular sampling. Figures \ref{fig:uneven} shows visual comparisons of PointCleanNet \cite{Marie2020PointCleanNet}, PointFilter \cite{2020Pointfilter}, and our method. We can see from Figure \ref{fig:uneven} that our method generates a very competitive result with preserving features and removing noise. 

\textbf{Runtime.}
We record the runtime of each step in our method and a total of 3 iterations using two different iteration schemes, shown in Table \ref{table:runtime_iteration}. As for the first iteration scheme, we observe that it takes a similar time for each step in each iteration. Often, step 1 involves more computation than step 2. With the second iteration scheme, since step 1 has been avoided in follow-up iterations, the total runtime over 3 iterations is much less than that of the first iteration scheme. 

We also record the runtime of some compared methods. For our method, the runtime of one iteration using two iteration schemes is recorded. As shown in Table \ref{table:runtime}, our method is slower than other methods. This can be attributed to the intensive computation involved in our method for finding and aggregating non-local useful information, i.e. in a non-local sense. By contrast, other methods only utilize local information, i.e. in a local sense. Note that our two different iteration schemes may induce slightly different runtime in the first iteration though they are the same in the first iteration. This is most likely due to the slight runtime discrepancy of the computer at that time. With regard to deep learning methods, the reported numbers are actually the reference phase time (i.e. without considering training time). In a nutshell, our method is more suitable for offline processing. 

\setlength{\tabcolsep}{3pt}
\begin{table*}[!h]
    \centering
    \caption{ Runtime performance (in seconds) for our method. }
    \label{table:runtime_iteration}
    \begin{tabular}{l | c | c c c | c c c | c c c | c }
        \hline
        \multirow{3}*{Models} & \multirow{3}*{\#. points} & \multicolumn{10}{c}{First Iteration Scheme}\\
        \cline{3-12}
        ~                     & ~                            & \multicolumn{3}{c|}{1st iteration} & \multicolumn{3}{c|}{2nd iteration} & \multicolumn{3}{c|}{3rd iteration} & \multirow{2}*{Total}\\
        \cline{3-11}
        ~                     & ~                            & Step 1 & Step 2 & Subtotal & Step 1 & Step 2 & Subtotal & Step 1 & Step 2 & Subtotal & ~ \\
        \hline
        Figure \ref{fig:synthetic} 1st row & 50419           & 1064     & 823      & 1887  & 1125     & 792      & 1917  & 1187     & 837      & 2024  & 5828 \\
        Figure \ref{fig:synthetic} 4th row & 63344           & 1254     & 344      & 1598  & 1220     & 339      & 1559  & 1291     & 324      & 1615  & 4772 \\
        Figure \ref{fig:synthetic} 7th row & 26519           & 505      & 434      & 939   & 517      & 538      & 1055  & 514      & 571      & 1085  & 3079 \\
        Figure \ref{fig:raw} 2nd row       & 97538           & 1909     & 895      & 2804  & 1929     & 762      & 2691  & 1894     & 803      & 2697  & 8192\\
        Figure \ref{fig:raw} 6th row       & 35920           & 701      & 1500     & 2201  & 686      & 1984     & 2670  & 685      & 2208     & 2893  & 7764 \\
        \hline
        \multirow{3}*{Models} & \multirow{3}*{\#. points} & \multicolumn{10}{c}{Second Iteration Scheme}\\
        \cline{3-12}
        ~                     & ~                            & \multicolumn{3}{c|}{1st iteration} & \multicolumn{3}{c|}{2nd iteration} & \multicolumn{3}{c|}{3rd iteration} & \multirow{2}*{Total} \\
        \cline{3-11}
        ~                     & ~                            & Step 1 & Step 2 & Subtotal & Step 1 & Step 2 & Subtotal & Step 1 & Step 2 & Subtotal & ~ \\
        \hline
        Figure \ref{fig:synthetic} 1st row & 50419           & 1067     & 909      & 1976  & -        & 817      & 817    & -       & 855      & 855   & 3648\\
        Figure \ref{fig:synthetic} 4th row & 63344           & 1227     & 328      & 1555  & -        & 307      & 307   & -        & 345      & 345   & 2207 \\
        Figure \ref{fig:synthetic} 7th row & 26519           & 517      & 459      & 976   & -        & 152      & 152   & -        & 173      & 173   & 1301 \\
        Figure \ref{fig:raw} 2nd row       & 97538           & 1924     & 780      & 2704  & -        & 635      & 635   & -        & 882      & 882   & 4221 \\
        Figure \ref{fig:raw} 6th row       & 35920           & 687      & 1500     & 2187  & -        & 98       & 98    & -        & 104      & 104   & 2389 \\
        \hline
    \end{tabular}
\end{table*}

\setlength{\tabcolsep}{3pt}
\begin{table}[htb]\tablefont
    \centering
    \caption{ Runtime performance (in seconds) for several compared methods. Note that * represents deep learning methods. }
    \label{table:runtime}
    \begin{tabular}{l c c c c c}
        \hline
        Methods & \tabincell{c}{Figure \ref{fig:synthetic}\\1st row} & \tabincell{c}{Figure \ref{fig:synthetic}\\4th row}  & \tabincell{c}{Figure \ref{fig:synthetic}\\7th row}  & \tabincell{c}{Figure \ref{fig:raw}\\2nd row} & \tabincell{c}{Figure \ref{fig:raw}\\6th row}\\
        \hline
        \small{PointCleanNet*} \cite{Marie2020PointCleanNet}    & 148  & 187   & 83   & 280   & 109\\
        PointFilter* \cite{2020Pointfilter}                     & 32   & 40    & 15   & 61    & 19\\
        CLOP \cite{preiner2014continuous}                       & 128  & 175   & 52   & 504   & 85 \\
        \hline
        Ours (Scheme 1)                          & 1887 & 1598  & 939  & 2804  & 2201  \\
        Ours (Scheme 2)                          & 1976 & 1555  & 976  & 2704  & 2187  \\
        \hline
    \end{tabular}
\end{table}

\subsection{Limitations and Future Work}
\begin{figure}[htb]
\centering
\begin{minipage}[b]{0.32\linewidth}
{\label{}\includegraphics[width=1\linewidth]{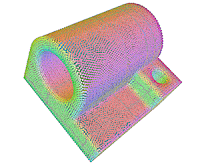}}
\centerline{Ground truth}
\end{minipage}
\begin{minipage}[b]{0.32\linewidth}
{\label{}\includegraphics[width=1\linewidth]{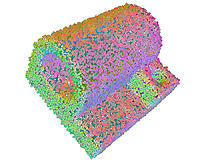}}
\centerline{Noisy input}
\end{minipage}
\begin{minipage}[b]{0.32\linewidth}
{\label{}\includegraphics[width=1\linewidth]{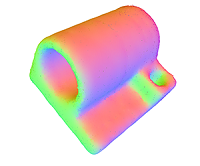}}
\centerline{Filtering result}
\end{minipage} 
\caption{ A failure example. Our method cannot well preserve sharp edges. }
\label{fig:limiataion}
\end{figure}

Although our method shows outstanding performance in point cloud filtering without taking account of normal information (see above sections), it still has few limitations. As discussed above, our method is slow due to the utilization of non-local similar information and is more suitable for offline applications. We plan to mitigate this with parallel computing in the future. 
As for the model with sharp edges (such as CAD models), our method might blur the edges while removing noise. An example is shown in Figure \ref{fig:limiataion}. This is mainly because that our method only utilizes the positional information, and averaging positions tends to over-smooth sharp edges as evidenced by previous research \cite{preiner2014continuous,huang2009consolidation}. Thanks to the non-local property, our method works very well in preserving sharp features in many other point clouds, as demonstrated in the above sections. In the future, we would like to exploit the possibilities of preserving sharp edges by using point positions only. 

\section{Conclusion}
\label{sec:conclusion}

We have presented a robust and effective approach for feature-preserving point cloud filtering, without requiring any normal information. It is a purely position based method. Given a noisy point cloud as input, our method first searches each local patch's similar patches, and then aggregate the non-local information (i.e. each central point) in a canonical space. Finally, we inversely map the aggregated coordinate to the original space. Our method is simple yet effective, and experimental results demonstrate that it generally outperforms state-of-the-art position based methods, and is comparable to or even better than normal based methods.

\section*{References}
\bibliography{mybibfile}

\begin{thebibliography}{10}
\expandafter\ifx\csname url\endcsname\relax
  \def\url#1{\texttt{#1}}\fi
\expandafter\ifx\csname urlprefix\endcsname\relax\def\urlprefix{URL }\fi
\expandafter\ifx\csname href\endcsname\relax
  \def\href#1#2{#2} \def\path#1{#1}\fi

\bibitem{lipman2007parameterization}
Y.~Lipman, D.~Cohen-Or, D.~Levin, H.~Tal-Ezer, Parameterization-free projection
  for geometry reconstruction, ACM Transactions on Graphics (TOG) 26~(3) (2007)
  22.

\bibitem{huang2009consolidation}
H.~Huang, D.~Li, H.~Zhang, U.~Ascher, D.~Cohen-Or, Consolidation of unorganized
  point clouds for surface reconstruction, ACM Transactions on Graphics (TOG)
  28~(5) (2009) 1--7.

\bibitem{preiner2014continuous}
R.~Preiner, O.~Mattausch, M.~Arikan, R.~Pajarola, M.~Wimmer, Continuous
  projection for fast l1 reconstruction, ACM Transactions on Graphics (TOG)
  33~(4) (2014) 1--13.

\bibitem{oztireli2009feature}
A.~C. {\"O}ztireli, G.~Guennebaud, M.~Gross, Feature preserving point set
  surfaces based on non-linear kernel regression, in: Computer Graphics Forum,
  Vol.~28, Wiley Online Library, 2009, pp. 493--501.

\bibitem{lu2017gpf}
X.~Lu, S.~Wu, H.~Chen, S.-K. Yeung, W.~Chen, M.~Zwicker, Gpf: Gmm-inspired
  feature-preserving point set filtering, IEEE transactions on visualization
  and computer graphics 24~(8) (2017) 2315--2326.

\bibitem{2020Pointfilter}
D.~Zhang, X.~Lu, H.~Qin, Y.~He, Pointfilter: Point cloud filtering via
  encoder-decoder modeling, IEEE Transactions on Visualization and Computer
  Graphics.

\bibitem{Marie2020PointCleanNet}
M.~Rakotosaona, V.~L. Barbera, P.~Guerrero, N.~J. Mitra, M.~Ovsjanikov,
  Pointcleannet: Learning to denoise and remove outliers from dense point
  clouds, Computer Graphics Forum 39~(1).

\bibitem{lu2020deep}
D.~Lu, X.~Lu, Y.~Sun, J.~Wang, Deep feature-preserving normal estimation for
  point cloud filtering, Computer-Aided Design 125 (2020) 102860.

\bibitem{levin1998approximation}
D.~Levin, The approximation power of moving least-squares, Mathematics of
  computation 67~(224) (1998) 1517--1531.

\bibitem{levin2004mesh}
D.~Levin, Mesh-independent surface interpolation, in: Geometric modeling for
  scientific visualization, Springer, 2004, pp. 37--49.

\bibitem{alexa2001point}
M.~Alexa, J.~Behr, D.~Cohen-Or, S.~Fleishman, D.~Levin, C.~T. Silva, Point set
  surfaces, in: Proceedings Visualization, 2001. VIS'01., IEEE, 2001, pp.
  21--29.

\bibitem{alexa2003computing}
M.~Alexa, J.~Behr, D.~Cohen-Or, S.~Fleishman, D.~Levin, C.~T. Silva, Computing
  and rendering point set surfaces, IEEE Transactions on visualization and
  computer graphics 9~(1) (2003) 3--15.

\bibitem{amenta2004defining}
N.~Amenta, Y.~J. Kil, Defining point-set surfaces, ACM Transactions on Graphics
  (TOG) 23~(3) (2004) 264--270.

\bibitem{fleishman2005robust}
S.~Fleishman, D.~Cohen-Or, C.~T. Silva, Robust moving least-squares fitting
  with sharp features, ACM Transactions on Graphics (TOG) 24~(3) (2005)
  544--552.

\bibitem{lange2005anisotropic}
C.~Lange, K.~Polthier, Anisotropic smoothing of point sets, Computer Aided
  Geometric Design 22~(7) (2005) 680--692.

\bibitem{liao2013efficient}
B.~Liao, C.~Xiao, L.~Jin, H.~Fu, Efficient feature-preserving local projection
  operator for geometry reconstruction, Computer-Aided Design 45~(5) (2013)
  861--874.

\bibitem{huang2013edge}
H.~Huang, S.~Wu, M.~Gong, D.~Cohen-Or, U.~Ascher, H.~Zhang, Edge-aware point
  set resampling, ACM Transactions on Graphics (TOG) 32~(1) (2013) 1--12.

\bibitem{avron2010}
H.~Avron, A.~Sharf, C.~Greif, D.~Cohen-Or, $\ell$1-sparse reconstruction of
  sharp point set surfaces, ACM Transactions on Graphics (TOG) 29~(5) (2010)
  1--12.

\bibitem{sun2015denoising}
Y.~Sun, S.~Schaefer, W.~Wang, Denoising point sets via l0 minimization,
  Computer Aided Geometric Design 35 (2015) 2--15.

\bibitem{yifan2019differentiable}
W.~Yifan, F.~Serena, S.~Wu, C.~{\"O}ztireli, O.~Sorkine-Hornung, Differentiable
  surface splatting for point-based geometry processing, ACM Transactions on
  Graphics (TOG) 38~(6) (2019) 1--14.

\bibitem{2020Lowrank}
X.~Lu, S.~Schaefer, J.~Luo, L.~Ma, Y.~He, Low rank matrix approximation for 3d
  geometry filtering, IEEE Transactions on Visualization and Computer Graphics.

\bibitem{liu2020feature}
Z.~Liu, X.~Xiao, S.~Zhong, W.~Wang, Y.~Li, L.~Zhang, Z.~Xie, A
  feature-preserving framework for point cloud denoising, Computer-Aided Design
  127 (2020) 102857.

\bibitem{dinesh2020point}
C.~Dinesh, G.~Cheung, I.~V. Baji{\'c}, Point cloud denoising via feature graph
  laplacian regularization, IEEE Transactions on Image Processing 29 (2020)
  4143--4158.

\bibitem{li2020dnf}
X.~Li, R.~Li, L.~Zhu, C.-W. Fu, P.-A. Heng, Dnf-net: A deep normal filtering
  network for mesh denoising, IEEE Transactions on Visualization and Computer
  Graphics.

\bibitem{roveri2018pointpronets}
R.~Roveri, A.~C. {\"O}ztireli, I.~Pandele, M.~Gross, Pointpronets:
  Consolidation of point clouds with convolutional neural networks, in:
  Computer Graphics Forum, Vol.~37, Wiley Online Library, 2018, pp. 87--99.

\bibitem{yu2018ec}
L.~Yu, X.~Li, C.-W. Fu, D.~Cohen-Or, P.-A. Heng, Ec-net: an edge-aware point
  set consolidation network, in: Proceedings of the European Conference on
  Computer Vision (ECCV), 2018, pp. 386--402.

\bibitem{yu2018pu}
L.~Yu, X.~Li, C.-W. Fu, D.~Cohen-Or, P.-A. Heng, Pu-net: Point cloud upsampling
  network, in: Proceedings of the IEEE Conference on Computer Vision and
  Pattern Recognition, 2018, pp. 2790--2799.

\bibitem{yin2018p2p}
K.~Yin, H.~Huang, D.~Cohen-Or, H.~Zhang, P2p-net: Bidirectional point
  displacement net for shape transform, ACM Transactions on Graphics (TOG)
  37~(4) (2018) 1--13.

\bibitem{hermosilla2019total}
P.~Hermosilla, T.~Ritschel, T.~Ropinski, Total denoising: Unsupervised learning
  of 3d point cloud cleaning, in: Proceedings of the IEEE/CVF International
  Conference on Computer Vision, 2019, pp. 52--60.

\bibitem{pistilli2020learning}
F.~Pistilli, G.~Fracastoro, D.~Valsesia, E.~Magli, Learning graph-convolutional
  representations for point cloud denoising, in: European Conference on
  Computer Vision, Springer, 2020, pp. 103--118.

\bibitem{luo2020differentiable}
S.~Luo, W.~Hu, Differentiable manifold reconstruction for point cloud
  denoising, in: Proceedings of the 28th ACM International Conference on
  Multimedia, 2020, pp. 1330--1338.

\bibitem{hu2020feature}
W.~Hu, X.~Gao, G.~Cheung, Z.~Guo, Feature graph learning for 3d point cloud
  denoising, IEEE Transactions on Signal Processing 68 (2020) 2841--2856.

\bibitem{zhu2017non}
L.~Zhu, C.-W. Fu, M.~S. Brown, P.-A. Heng, A non-local low-rank framework for
  ultrasound speckle reduction, in: Proceedings of the IEEE Conference on
  Computer Vision and Pattern Recognition, 2017, pp. 5650--5658.

\bibitem{zhu2016non}
L.~Zhu, C.-W. Fu, Y.~Jin, M.~Wei, J.~Qin, P.-A. Heng, Non-local sparse and
  low-rank regularization for structure-preserving image smoothing, in:
  Computer Graphics Forum, Vol.~35, Wiley Online Library, 2016, pp. 217--226.

\bibitem{li2018non}
X.~Li, L.~Zhu, C.-W. Fu, P.-A. Heng, Non-local low-rank normal filtering for
  mesh denoising, in: Computer Graphics Forum, Vol.~37, Wiley Online Library,
  2018, pp. 155--166.

\bibitem{deschaud2010point}
J.-E. Deschaud, F.~Goulette, Point cloud non local denoising using local
  surface descriptor similarity, IAPRS 38~(3A) (2010) 109--114.

\bibitem{digne2012similarity}
J.~Digne, Similarity based filtering of point clouds, in: 2012 IEEE computer
  society conference on computer vision and pattern recognition workshops,
  IEEE, 2012, pp. 73--79.

\bibitem{rosman2013patch}
G.~Rosman, A.~Dubrovina, R.~Kimmel, Patch-collaborative spectral point-cloud
  denoising, in: Computer Graphics Forum, Vol.~32, Wiley Online Library, 2013,
  pp. 1--12.

\bibitem{bm3d}
K.~Dabov, A.~Foi, V.~Katkovnik, K.~Egiazarian, Image denoising by sparse 3-d
  transform-domain collaborative filtering, IEEE Transactions on image
  processing 16~(8) (2007) 2080--2095.

\bibitem{digne2017sparse}
J.~Digne, S.~Valette, R.~Chaine, Sparse geometric representation through local
  shape probing, IEEE transactions on visualization and computer graphics
  24~(7) (2017) 2238--2250.

\bibitem{sarkar2018structured}
K.~Sarkar, F.~Bernard, K.~Varanasi, C.~Theobalt, D.~Stricker, Structured
  low-rank matrix factorization for point-cloud denoising, in: 2018
  International Conference on 3D Vision (3DV), IEEE, 2018, pp. 444--453.

\bibitem{zeng20193d}
J.~Zeng, G.~Cheung, M.~Ng, J.~Pang, C.~Yang, 3d point cloud denoising using
  graph laplacian regularization of a low dimensional manifold model, IEEE
  Transactions on Image Processing 29 (2019) 3474--3489.

\bibitem{wright2009robust}
J.~Wright, A.~Ganesh, S.~Rao, Y.~Peng, Y.~Ma, Robust principal component
  analysis: Exact recovery of corrupted low-rank matrices via convex
  optimization, in: Advances in neural information processing systems, 2009,
  pp. 2080--2088.

\bibitem{bouwmans2014robust}
T.~Bouwmans, E.~H. Zahzah, Robust pca via principal component pursuit: A review
  for a comparative evaluation in video surveillance, Computer Vision and Image
  Understanding 122 (2014) 22--34.

\bibitem{sobral2016lrslibrary}
A.~Sobral, T.~Bouwmans, E.-h. Zahzah, Lrslibrary: Low-rank and sparse tools for
  background modeling and subtraction in videos, Robust Low-Rank and Sparse
  Matrix Decomposition: Applications in Image and Video Processing.

\bibitem{lau2019restoration}
C.~P. Lau, Y.~H. Lai, L.~M. Lui, Restoration of atmospheric
  turbulence-distorted images via rpca and quasiconformal maps, Inverse
  Problems 35~(7) (2019) 074002.

\bibitem{chen2017denoising}
Y.~Chen, Y.~Guo, Y.~Wang, D.~Wang, C.~Peng, G.~He, Denoising of hyperspectral
  images using nonconvex low rank matrix approximation, IEEE Transactions on
  Geoscience and Remote Sensing 55~(9) (2017) 5366--5380.

\bibitem{mattei2017point}
E.~Mattei, A.~Castrodad, Point cloud denoising via moving rpca, in: Computer
  Graphics Forum, Vol.~36, Wiley Online Library, 2017, pp. 123--137.

\bibitem{chen2019multi}
H.~Chen, M.~Wei, Y.~Sun, X.~Xie, J.~Wang, Multi-patch collaborative point cloud
  denoising via low-rank recovery with graph constraint, IEEE transactions on
  visualization and computer graphics 26~(11) (2019) 3255--3270.

\bibitem{2009Rank}
V.~Chandrasekaran, S.~Sanghavi, P.~A. Parrilo, A.~S. Willsky, Rank-sparsity
  incoherence for matrix decomposition, SIAM Journal on Optimization 21~(2)
  (2009) 572--596.

\bibitem{ma2018efficient}
S.~Ma, N.~S. Aybat, Efficient optimization algorithms for robust principal
  component analysis and its variants, Proceedings of the IEEE 106~(8) (2018)
  1411--1426.

\bibitem{Lin2010The}
Z.~Lin, M.~Chen, Y.~Ma, The augmented lagrange multiplier method for exact
  recovery of corrupted low-rank matrices, Mathematical Programming 9.

\bibitem{Emmanuel2009Exact}
E.~J. Candès, B.~Recht, Exact matrix completion via convex optimization,
  Foundations of Computational Mathematics 9~(6) (2009) 717.

\bibitem{yuan2009sparse}
X.~Yuan, J.~Yang, Sparse and low rank matrix decomposition via alternating
  direction method, Pacific Journal of Optimization 9.

\bibitem{2009Robust_Book}
E.~J. Cand{\`e}s, X.~Li, Y.~Ma, J.~Wright, Robust principal component analysis,
  Journal of the ACM (JACM) 58~(3) (2011) 1--37.

\end{thebibliography}

\end{document}